\documentclass[sn-basic,iicol]{sn-jnl}% Default with double column layout

\usepackage{graphicx}%
\usepackage{multirow} 
\usepackage{amsmath,amssymb,amsfonts}%
\usepackage{amsthm}%
\usepackage{mathrsfs}%
\usepackage[title]{appendix}%
\usepackage{xcolor}%
\usepackage{textcomp}%
\usepackage{manyfoot}%
\usepackage{booktabs}%
\usepackage{algorithm}%
\usepackage{algorithmicx}%
\usepackage{algpseudocode}%
\usepackage{listings}%
\usepackage{makecell}

\newcommand{\tablestyle}[2]{\setlength{\tabcolsep}{#1}\renewcommand{\arraystretch}
{#2}\centering\footnotesize}

\usepackage{color}

\begin{document}

\title[Article Title]{
A Survey on Deep Clustering: From the Prior Perspective
}

\author{\fnm{Yiding} \sur{Lu}}\email{yidinglu.gm@gmail.com}
\equalcont{These authors contributed equally to this work.}

\author{\fnm{Haobin} \sur{Li}}\email{haobinli.gm@gmail.com}
\equalcont{These authors contributed equally to this work.}

\author{\fnm{Yunfan} \sur{Li}}\email{yunfanli.gm@gmail.com}

\author{\fnm{Yijie} \sur{Lin}}\email{linyijie.gm@gmail.com}

\author*{\fnm{Xi} \sur{Peng\textsuperscript{*}}}\email{pengx.gm@gmail.com}

\affil{\orgdiv{College of Computer Science}, \orgname{Sichuan University}, \orgaddress{\city{Chengdu}, \state{Sichuan}, \country{China}}}

\abstract{Facilitated by the powerful feature extraction ability of neural networks, deep clustering has achieved great success in analyzing high-dimensional and complex real-world data. The performance of deep clustering methods is affected by various factors such as network structures and learning objectives. However, as pointed out in this survey, the essence of deep clustering lies in the incorporation and utilization of prior knowledge, which is largely ignored by existing works. From pioneering deep clustering methods based on data structure assumptions to recent contrastive clustering methods based on data augmentation invariances, the development of deep clustering intrinsically corresponds to the evolution of prior knowledge. In this survey, we provide a comprehensive review of deep clustering methods by categorizing them into six types of prior knowledge. We find that in general the prior innovation follows two trends, namely, \textit{i)} from mining to constructing, and \textit{ii)} from internal to external. 
Besides, we provide a benchmark on five widely-used datasets and analyze the performance of methods with diverse priors. By providing a novel prior knowledge perspective, we hope this survey could provide some novel insights and inspire future research in the deep clustering community.}


\maketitle

\section{Introduction}

As a fundamental problem in machine learning, clustering aims at grouping data instances into several clusters, where instances from the same cluster share similar semantics and instances from different clusters are dissimilar. Clustering could reveal the inherent semantic structure underlying the data, which benefits the down-stream analysis such as anomaly detection~\citep{anomalydetection2018}, person re-identification~\citep{reid2021}, community detection~\citep{communitydetection}, and domain adaption~\citep{uda2023}, etc.

In the early stage, various classic clustering methods are developed, such as centroid-based clustering \citep{kmeans1967}, density-based clustering \citep{DBSCAN1996}, hierarchical clustering \citep{hierarchical2012}, and so on. These shallow methods are grounded in theory and enjoy high interpretability. Later on, some works extend shallow clustering methods to diverse data types, such as multi-view \citep{zhang2015,nie2016parameter, mvc_graph2017,wang2018} and graph data \citep{newman2004finding, schaeffer2007graph}. Other efforts have been made to improve the scalability \citep{zhang2023large} of shallow clustering methods.

However, shallow clustering methods partition instances based on the similarity~\citep{kmeans1967} or density~\citep{DBSCAN1996} of the given raw or linear transformed data. Due to the limited feature extraction ability, shallow clustering methods would achieve sub-optimal results when confronted with complex, high-dimensional, and non-linear data in the real world. To tackle this challenge, deep clustering techniques are proposed to incorporate neural networks into clustering methods. In other words, deep clustering simultaneously learns discriminative representations and performs clustering on the learned features, progressively benefiting each other. 

Over the past few years, many efforts have been devoted to improving the clustering performance from various aspects, such as network architectures~\citep{DeepCluster2018, vit_cluster2021}, training strategies~\citep{ClusterGAN2019}, and loss functions~\citep{VaDE2016, GCC2021}. However, we would like to highlight that the fundamental challenge of deep clustering is the absence of data annotations. Consequently, the key to deep clustering lies in introducing proper \textbf{prior knowledge} to construct the supervision signals. From the early data structure assumption to the recent data augmentation invariance, the development of deep clustering methods intrinsically corresponds to the evolution of prior knowledge. In this survey, we provide a comprehensive review of deep clustering methods from the perspective of prior knowledge.

Inspired by traditional clustering and dimensionality reduction approaches~\citep{LLE2000, LEcluster2001}, the early deep clustering methods~\citep{DEN2014, PARTY2016, SpectralNet2018} build upon the \textbf{structure prior} of data. Based on the assumption that the inherent data structure could reflect the semantic relation, these methods incorporate classic manifold~\citep{LLE2000} or subspace learning~\citep{SparseRepresentaion2010} objectives to optimize the neural network for feature extraction and clustering. The second type of prior knowledge is the \textbf{distribution prior}, which assumes that instances from different clusters follow distinct distributions. Based on such a prior, several generative deep clustering methods~\citep{VaDE2016, ClusterGAN2019} propose to learn the latent distribution of samples for the data partition. In the past few years, the success of contrastive learning spawns a new category of prior knowledge, namely, \textbf{augmentations invariance}. Instead of mining data priors, researchers turn to constructing additional priors with the data augmentation technique. Leveraging the invariance across different augmented samples at the instance representation and clustering assignment levels, numerous contrastive clustering methods~\citep{IIC2019, CC2021} significantly improve the feature discriminability and clustering performance. Further, researchers find that instances of the same semantics are likely to be mapped into nearby points in the latent space, and accordingly propose the \textbf{neighborhood consistency} prior. Specifically, by encouraging neighboring samples to have similar cluster assignments, several works~\citep{SCAN2020, GCC2021} alleviate the false-negative problem in the contrastive clustering paradigm, thus advancing the clustering results. Another branch of progress is made based on the \textbf{pseudo label} prior, namely, cluster assignments with high confidence are likely to be correct. By selecting confident predictions as pseudo labels, several studies further boost the clustering performance through pseudo-labeling~\citep{TCL2022, SeCu2023} and semi-supervised learning~\citep{SPICE2022}. Very recently, instead of pursuing internal priors from the data itself, some works~\citep{SIC2023, TAC2023} attempt to introduce abundant \textbf{external knowledge} such as textual descriptions to guide clustering.

In summary, the essence of deep clustering lies in how to find and leverage effective prior knowledge, for both feature extraction and cluster assignment. To provide an overview of the development of deep clustering, in this paper, we categorize a series of state-of-the-art approaches according to the taxonomy of prior knowledge. We hope such a new perspective for deep clustering could inspire future research in the community. The rest of this paper is organized as follows: First, Section~\ref{sec:prob_def} introduces the preliminaries on deep clustering. Section~\ref{sec:priors} reviews the existing deep clustering methods from the prior knowledge perspective. Then, Section~\ref{sec:expmt} provides experimental analyses of deep clustering methods. After that, Section~\ref{sec:appli} briefly introduces some applications of deep clustering in the vicinagearth security. Lastly, Section~\ref{sec:chall} summarizes some notable trends and challenges for deep clustering.

\newpage
\subsection*{Related Surveys}
We notice that several surveys on deep clustering have been proposed in recent years. Briefly, \citet{SurveyOnArch2018} categorizes deep clustering methods according to the network architecture. \citet{SurveyOnApp2022} focuses on applications of deep clustering. \citet{SurveyOnData2022} summarizes existing methods from the view of data types, such as single- and multi-view. \citet{SurveyOnTrain2022} discusses various interactions between representation learning and clustering. Distinct from existing surveys, this work systematically provides a new perspective from the prior knowledge, which plays a more intrinsic and essential role in deep clustering.

\section{Problem Definition}
\label{sec:prob_def}
In this section, we introduce the pipeline of deep clustering, including the notation and problem definition. Unless specially notified, in this paper, we use bold uppercase and lowercase to denote matrices and vectors, respectively. The commonly used notations are summarized in Table~\ref{tabel:notations}.

The deep clustering problem is formally defined as follows: given a set of instances $\mathcal{D}=\left\{\mathbf{x}_i\right\}_{i=1}^{N}\in \mathcal{X}$ that belongs to $C$ classes, deep clustering aims to learn discriminative features and group the instances into $C$ clusters according to their semantics. Specifically, deep clustering methods first learn a deep neural network $f:\mathcal{X}\rightarrow \mathcal{Z}$ for feature extraction, \textit{i.e.}, $\mathbf{z}_i=f(\mathbf{x}_i)$. Given instance features in the latent space, clustering results could be obtained in two ways. The most straightforward way is to apply classic algorithms such as K-means~\citep{kmeans1967} and DBSCAN~\citep{DBSCAN1996} on the learned features. The other solution is to train an additional cluster head $h:\mathcal{Z}\rightarrow \mathbb{R}^{C}$ to produce soft cluster assignment $\mathbf{p}_i=\operatorname{softmax}(h(\mathbf{z}_i))$ which satisfies $\sum_{j=0}^{K}\mathbf{p}_{ij}=1$. The hard cluster assignment for the $i$-th instance could be computed by $\arg\max$ operation, namely,
\begin{equation}
    \tilde{y}_i=\arg\max_j\ \mathbf{p}_{ij}, 1\leq j\leq C.
\end{equation}
The cluster assignments provide the inherent semantic structure underlying the data, which could be utilized in various downstream analyses.

\begin{figure*}[t]
   \centering
   \includegraphics[width=0.95\textwidth]{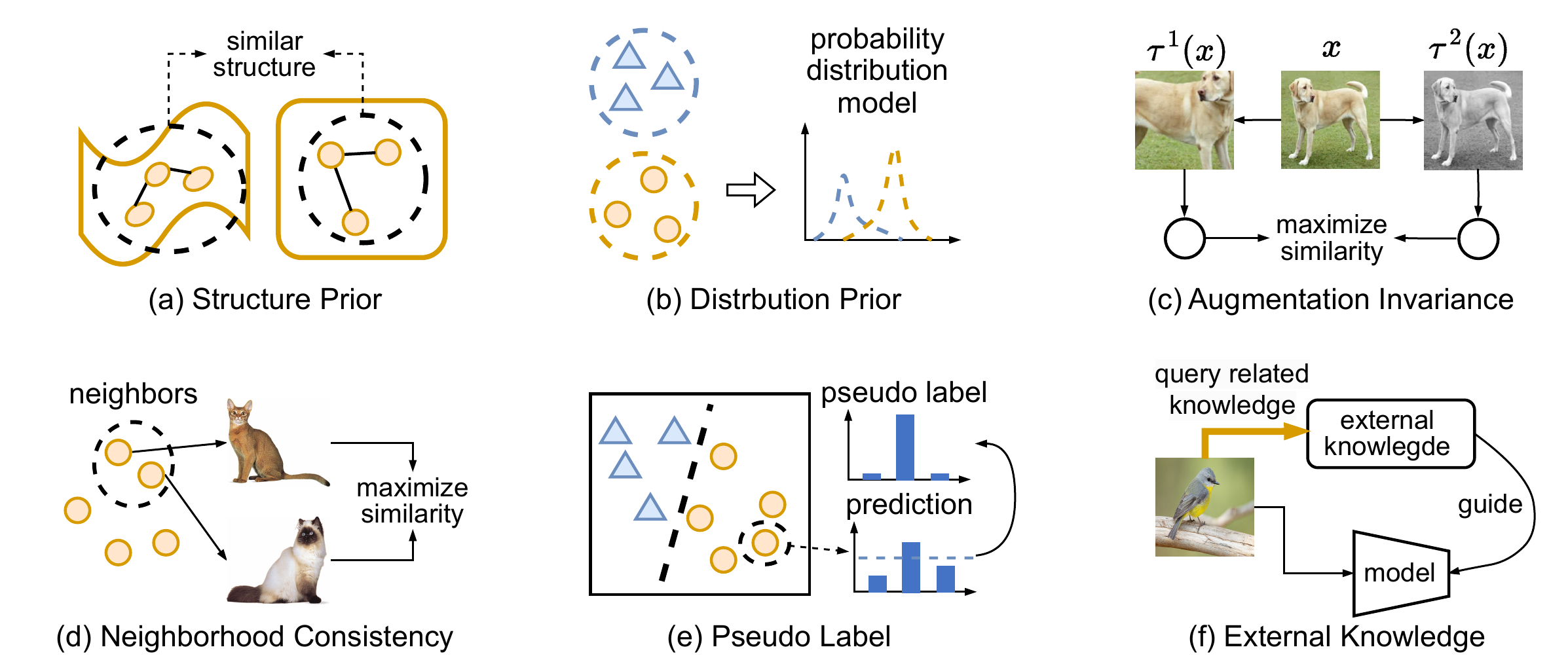}
   \caption{Six categories of prior knowledge for deep clustering. (a) Structure Prior: data structure could reflect the semantic relation between instances. (b) Distribution Prior: instances from different clusters follow distinct data distributions. (c) Augmentation Invariance: samples augmented by the same instance have similar features. (d) Neighborhood Consistency: neighboring samples have consistent cluster assignments. (e) Pseudo Label: cluster assignments with high confidence are likely to be correct. (f) External Knowledge: abundant knowledge favorable to clustering exists in open-world data and models.}
   \label{fig: prior}
\end{figure*}

\begin{table}[t]
\caption{Commonly used mathematical notations.}
\centering
\tablestyle{6pt}{1.4}{
\begin{tabular}{cl}
\toprule
Notation       & Explanation                           \\ \midrule
$N$            & Number of data instances           \\
$B$            & Size of a mini-batch \\
$C$            & Number of clusters                 \\
$f(\cdot)$     & Encoder network    \\
$h(\cdot)$     & Cluster head   \\
$\mathbf{x}_i$ & $i$-th data instance               \\
$\mathbf{z}_i$ & Feature of the $i$-th instance  \\
$\tilde{y}_i$  & Pseudo label of the $i$-th instance \\
$\|\cdot\|$    & L2-norm of a vector                    \\
$\langle\cdot\rangle$    & Dot product operator       \\
$\operatorname{s}(\mathbf{a},\mathbf{b})$ & Cosine similarity, \textit{i.e.}, $\operatorname{s}(\mathbf{a},\mathbf{b})=\frac{\langle\mathbf{a},\mathbf{b}\rangle}{\|\mathbf{a}\|\|\mathbf{b}\|}$ \\
$\mathbf{c}_i$ & Centroid of the $i$-th cluster        \\

$H(\cdot)$ & Entropy, \textit{i.e.}, $H(X)=\sum_{x\in X}-p(x)\log p(x)$ \\
$H(\cdot,\cdot)$ & \makecell[l]{\\Conditional entropy,\textit{i.e.},\\$H(Y\mid X)=\sum_{x\in X,y\in Y}-p(x,y)\log\frac{p(x,y)}{p(x)}$}\\
$I(\cdot;\cdot)$ & \makecell[l]{\\Mutual Information, \textit{i.e.}, \\$I(X;Y)=H(X)-H(X\mid Y)$}\\
$\tau$ & Temperature coefficient of contrastive loss\\
\bottomrule
\end{tabular}}
\label{tabel:notations}
\end{table}

\section{Priors for Deep Clustering}
\label{sec:priors}
In this section, we review existing deep clustering methods from the perspective of prior knowledge. The priors are illustrated in Figure~\ref{fig: prior} and the method categorization is summarized in Table\ref{tab:summary}.

\subsection{Structure Prior}
Structure prior is mostly inspired by traditional clustering methods. Traditional cluster is mainly rooted in assumptions about the structural characteristics of clusters in data space. For example, K-means~\citep{kmeans1967} aims to learn $k$ cluster centroids, which assumes that instances in each cluster form a spherical structure around its centroid. DBSCAN~\citep{DBSCAN1996} is based on the assumption that a cluster in data space is a contiguous region of high point density, separated from other such clusters by regions of low point density. Spectral clustering~\citep{LEcluster2001} assumes data lies on a locally linear manifold so that the local neighborhoods' relation should be preserved in latent space. Those methods partition instances according to the graph Laplacian. Agglomerative clustering~\citep{AggCluster1978} considers the hierarchical structure of data and performs clustering with merging and splitting. Motivated by the success of classic clustering methods, the early exploration of deep clustering mainly focuses on adapting mature structure priors as objective functions to optimize neural networks.

Given well-structured data in the latent space, ABDC~\citep{ABDC2013} iteratively optimizes the data representation and clustering centers motivated by K-means. As the deep extension of classic spectral clustering, DEN~\citep{DEN2014}, SpectralNet~\citep{SpectralNet2018}, and MvLNet~\citep{huang2019mvscn,huang2021deep} compute the graph Laplacian in the latent space learned by auto-encoder~\citep{AE2013} and SiameseNets~\citep{DrLIM2006, SiameseNets2018}, respectively. Likewise, DCC~\citep{DCC2018} extends the core idea of RCC~\citep{RCC2017} by performing a relation matching based on the similarity between latent features. The auto-encoder is then optimized by minimizing the distance of paired instances in the latent space. PARTY~\citep{PARTY2016} is the first deep subspace clustering method, which introduces the sparsity prior and self-representation property in subspace learning to optimize neural networks. Motivated by the hierarchical structure of clusters, JULE~\citep{JULE2016} achieves agglomerative deep clustering by progressively merging clusters and optimizing the features.

\subsection{Distribution Prior}
Distribution prior refers to instances of different semantics following distinct data distributions. Such a prior arouses the generative deep clustering paradigm, which employs variational autoencoder~\citep{VAE2013} (VAE) and generative adversarial network~\citep{GAN2014} (GAN) to learn the underlying distribution. Instances generated from similar distributions are then grouped together to achieve clustering.

VaDE~\citep{VaDE2016} is the first deep generative clustering method, which computes different data distributions by fitting the Gaussian mixture model (GMM) in the latent space. To generate an instance, VaDE first samples a cluster distribution $p\left(c\right)$ to generate a latent vector $p\left(z\mid c\right)$, and then reconstructs the instance in the input space $p\left(x \mid z\right)$. The cluster assignment and neural network are jointly optimized by maximizing the log-likelihood of instance, \textit{i.e.},
\begin{equation}
    \log p(\mathrm{x})=\log \int_{\mathrm{z}} \sum_c p(\mathrm{x} \mid \mathrm{z}) p(\mathrm{z} \mid c) p(c) \mathrm{dz}.
    \label{Eq: VaDE_1}
\end{equation}
Since directly computing Eq.~\ref{Eq: VaDE_1} is intractable, the optimization is approximated by the evidence lower bound (ELBO) of variational inference objective, namely,
\begin{equation}
    \mathcal{L}=\mathbb{E}_{q(\mathrm{z}, c \mid \mathrm{x})}\left[\log \frac{p(\mathrm{x}, \mathrm{z}, c)}{q(\mathrm{z}, c \mid \mathrm{x})}\right],
    \label{Eq: VaDE_2}
\end{equation}
where $q(\mathrm{z}, c\mid\mathrm{x})$ is variational posterior, which approximates the real posterior. The reparameterization trick introduced in VAE~\citep{VAE2013} is adopted to make the sampling process differentiable.

\begin{figure}[t]
   \centering
   \includegraphics[width=\columnwidth]{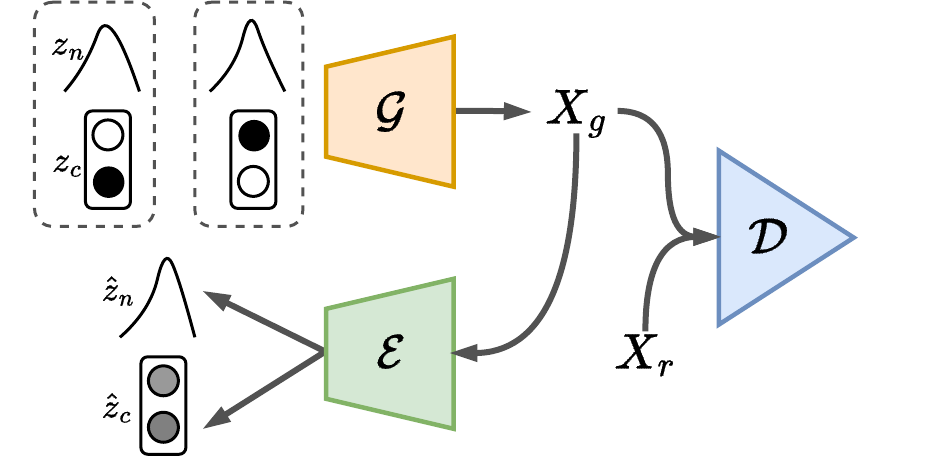}
   \caption{The framework of distribution prior based methods. In addition to the standard continuous latent variable $\mathbf{z}_n$, generative deep clustering methods further introduce a discrete variable $\mathbf{z}_c$ to capture the cluster information.
    }
   \label{fig: distrbution_prior}
\end{figure}

Though GMM could effectively distinguish distributions, Gaussian components are proved to be redundant, which harms the discriminability between different clusters~\citep{Deligan2017}. As an improvement, ClusterGAN, DCGAN~\citep{ClusterGAN2019,DCGAN2016} proposes to adopt GAN to implicitly learn the latent distributions. Specifically, in addition to the continuous latent variable $\mathbf{z}_n$, it introduces a one-hot encoding $\mathbf{z}_c$ to capture cluster distribution during the generation. The objective function of ClusterGAN is formulated as follows:
\begin{equation}
\begin{aligned}
\mathcal{L}=& \underset{\mathbf{x} \sim p_X(\mathbf{x})}{\mathbb{E}} q(\mathcal{D}(\mathbf{x}))+\underset{\mathbf{z} \sim \mathbb{P}_\mathbf{z}}{\mathbb{E}} q(1-\mathcal{D}(\mathcal{G}(\mathbf{z}))) \\
& \quad+\beta_n \underset{p_{\mathcal{Z}}(\mathbf{z})}{\mathbb{E}}\left\|\mathbf{z}_n-\mathcal{E}\left(\mathcal{G}\left(\mathbf{z}_n\right)\right)\right\|_2^2\\
&\quad+\beta_c \underset{p_{\mathcal{Z}}(\mathbf{z})}{\mathbb{E}} \mathcal{H}\left(\mathbf{z}_c, \mathcal{E}\left(\mathcal{G}\left(\mathbf{z}_c\right)\right)\right),
\end{aligned}
\end{equation}
where $\mathbf{z}=(\mathbf{z}_n,\mathbf{z}_c)$ is the mixed latent variable, $\mathcal{E}$ is the inverse network which maps data from the raw to latent space, $\mathcal{H}\left(\cdot,\cdot \right)$ denotes the cross-entropy, and $\beta_n$, $\beta_c$ are the weight parameters. The first two terms are consistent with standard GAN. The last two clustering-specific terms encourage a more distinct cluster distribution, as well as map inputs to the latent space to achieve clustering.

\begin{figure}[t]
   \centering
   \includegraphics[width=\columnwidth]{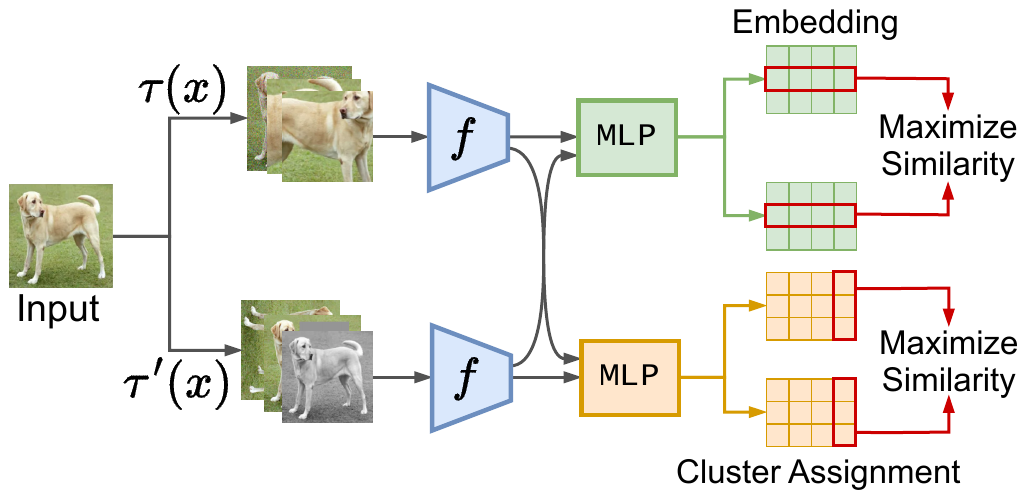}
   \caption{The framework of augmentation invariance based methods. Diverse transformations are first applied to augment the input data $x$, after which the shared deep neural network is utilized to extract features. The augmented samples of the same instance are encouraged to have similar features and cluster assignments.
}
   \label{fig: Augmentation_Invariance}
\end{figure}
\subsection{Augmentation Invariance}
In recent years, image augmentation methods~\citep{augmentation2019} have gained widespread attention, grounded in the prior that augmentations of the same instance could preserve consistent semantic information. 
This augmentation-invariance character inspires exploration of how to leverage the positive pairs (\textit{i.e.,} different augmentations of the same image) with similar semantic information. Notably, mutual-information-based methods and contrastive-learning-based methods have emerged as pioneers in the realm of deep clustering. In this section, we delve into the fundamental concepts and related works of both mutual-information-based and contrastive-learning-based methods.

Firstly, mutual information is a measure of dependence between two continuous random variables $X$ and $Y$, formally,
\begin{equation}
I(X ; Y)=\int_Y \int_X p(x, y) \log \left(\frac{p(x, y)}{p(x) p(y)}\right) d x d y,
\label{eq: MI}
\end{equation}
where $p(x, y)$ is the joint probability mass function of $X$ and $Y$, $p(x)$ and $p(y)$ are the marginal probability mass functions of $X$ and $Y$ respectively. In the context of information theory, leveraging the mutual information between variables of positive instances could enhance the optimization of clustering-related information.

IMSAT~\citep{IMSAT2017} stands as a typical information-theoretic approach to deep clustering. Its fundamental concept includes enforcing invariance on pair-wise augmented instances and achieving unambiguous and uniform cluster assignments. Specifically, IMSAT encourages the representations of augmented instances to closely match the representations of the original instances, \textit{i.e.}, 
\begin{equation}
\begin{aligned}
    \mathcal{L}=-\sum_{i, k} \mathbf{p}_{i k} \log \mathbf{p}^{\prime}_{i k}
\end{aligned}
\end{equation}
where $\mathbf{p}^{\prime}$ is the prediction representations of augmented instances.
This aspect can be viewed as exploring the maximization of mutual information between data and its augmentations. 
Besides, IMSAT implements regularized information maximization for deep clustering inspired by RIM~\citep{RIM2010} to keep the cluster assignments unambiguous and uniform. Specifically, IMSAT seeks to maximize the mutual information between instances and their cluster assignments, expressed as:
\begin{equation}
\begin{aligned}
    I(X ; Y)&= H(Y)-H(Y \mid X)\\
    &=-\sum_k \mathbf{p}_{\cdot k} \log \mathbf{p}_{\cdot k}+\frac{1}{N} \sum_{i, k} \mathbf{p}_{i k} \log \mathbf{p}_{i k},
\end{aligned}
\label{eq: IMSAT_1}
\end{equation}
where $H(\cdot)$ and $H(\cdot|\cdot)$ the entropy and conditional entropy, and $\mathbf{p}_{\cdot k}=\frac{1}{N} \sum_{i} \mathbf{p}_{i k}$. Increasing the first term (marginal entropy $H(Y)$) encourages uniform cluster assignments, \textit{i.e.}, the number of instances in each cluster tends to be the same. Conversely, decreasing the second term (conditional entropy $H(Y\mid X)$) encourages each instance to be unambiguously assigned to a certain cluster.  

IIC~\citep{IIC2019} and Completer~\citep{Completer2021,lin2022dcp} take a further step in exploring the mutual information between instances and their augmentations. The fundamental concept is to maximize the mutual information between the cluster assignments of pair-wise augmented instances. Specifically, IIC achieves semantically meaningful clustering and avoids trivial solutions by maximizing the mutual information between the cluster assignments,
\begin{equation}
\begin{aligned}
    \mathcal{L}&=I\left(Z, Z^{\prime}\right)=\sum_i^N I\left(\mathbf{z}_i, \mathbf{z}_i^{\prime}\right)=I(\mathbf{P}),\\
    &=\sum_{c=1}^C \sum_{c^{\prime}=1}^C \mathbf{P}_{c c^{\prime}} \cdot \ln \frac{\mathbf{P}_{c c^{\prime}}}{\mathbf{P}_c \cdot \mathbf{P}_{c^{\prime}}},
\end{aligned}
\label{eq: IIC_1}
\end{equation}
where $\mathbf{z}$ and $\mathbf{z}^{\prime}$ are the representations of the original instance $x$ and its augmentation $\mathbf{x}^{\prime}$, respectively. 
The conditional joint distribution of $\mathbf{z}$ and $\mathbf{z}^{\prime}$ is given by the matrix $\mathbf{P} \in \mathbb{R}^{C \times C}$ which is constituted by,
\begin{equation}
    \mathbf{P}=\frac{1}{n} \sum_{i=1}^n \mathbf{z}_i \cdot \left(\mathbf{z}_i^{\prime}\right)^{\top},
\end{equation}
where $\mathbf{P}_{c c^{\prime}}=P\left(z=c, z^{\prime}=c^{\prime}\right)$ denotes the element of $c$-th row and $c^{\prime}$-th column. Additionally, the marginals $\mathbf{P}_c=P(z=c)$ and $\mathbf{P}_{c^{\prime}}=P\left(z^{\prime}=c^{\prime}\right)$ can be obtained by summing over the rows and columns of this matrix. Notably, IIC stands out as one of the earliest deep frameworks designed entirely under the framework of information theory, distinguishing itself from IMSAT.

Similar to mutual-information-based methods, contrastive-learning-based methods treat instances augmented from the same instance as positive samples and the rest as negative samples. Let $\mathbf{z}_{2i}$ and $\mathbf{z}_{2i-1}$ represent two augmented representation of the $i$-th instance, the contrastive loss is formulated as:
\begin{equation}
\begin{aligned}
    \mathcal{L}&=\sum_i^N\left(\ell\left(2i,2i-1\right)+\ell\left(2i-1,2i\right)\right),\\
    \ell \left(i,j\right)&=-\log \frac{\exp\left(\operatorname{s} \left(\mathbf{z}_{i} \cdot \mathbf{z}_{j}\right) / \tau\right)}{\sum_{j=1}^{2 N} \mathbf{1}_{[j \neq i]} \exp\left(\operatorname{s} \left(\mathbf{z}_{i} \cdot \mathbf{z}_{j} / \right)\tau\right)},
\end{aligned}
\label{eq: con_1}
\end{equation}
where $\ell\left(i, j\right)$ represents the pairwise contrastive loss and $\tau$ controls the temperature of the softmax. The function $\operatorname{s} \left(\mathbf{z}_{i}, \mathbf{z}_{j}\right)$ denotes the similarity between representations $\mathbf{z}_{i}$ and $\mathbf{z}_{j}$. This loss encourages representations of positive instances to be closer while being separated from negative instances, encouraging meaningful clustering patterns. 

Notably, some theoretical works~\citep{CPC2018,ContraQuad2022,GraphM2023} have demonstrated that contrastive learning is equivalent to maximizing the mutual information from the instance level. Motivated by this observation, researchers have further explored the application of contrastive loss at the cluster level, proving beneficial for deep clustering. PICA~\citep{PICA2020} is one of the pioneer works of this domain. 
The fundamental concept behind it is to maximize the similarity between the cluster assignment of original and augmented data. This objective can be likened to conducting contrastive learning~\citep{liu2022improve} at the cluster level.
Motivated by PICA, CC~\citep{CC2021} and DRC~\citep{DRC2020} conduct contrastive learning on both instance level and cluster level. Specifically, cluster-level contrastive loss helps learn discriminative cluster assignment, which is the key to the clustering task. Formally, the cluster-level contrastive loss is,
\begin{equation}\begin{small}
\begin{aligned}
    \mathcal{L}&=\frac{1}{2 C} \sum_{i=1}^C\left(\ell\left(2i-1,2i\right)\! +\! \ell\left(2i,2i-1\right)\right)\! -\! H(\mathbf{Y}),\\
    \ell\left(i,j\right)&=-\log \frac{\exp \left(s\left(\mathbf{y}_i, \mathbf{y}_i\right) / \tau\right)}{\sum_{j=1}^{2 C} \mathbf{1}_{[j \neq i]}\left[\exp \left(s\left(\mathbf{y}_{i}, \mathbf{y}_j\right) / \tau\right)\right]},
\end{aligned}
\label{Eq: cc}\end{small}
\end{equation}
where $\mathbf{y}_i \in \mathbb{R}^{1\times N}$ is the cluster-level assignment and $\tau$ is the cluster-level temperature parameter. $H(\mathbf{Y}) = H(\mathbf{Y}^1)+H(\mathbf{Y}^2)$ is the cluster assignment probabilities entropy of two augmentations. The inclusion of $H(\mathbf{Y})$ helps avoid the trivial solution where most instances are assigned to the same cluster. Notably, the utilization of contrastive learning at the cluster level in CC and DRC has inspired subsequent works in the field.

TCC~\citep{TCC2021} takes a step further in exploring the interaction between instance-level and cluster-level representations. The core idea is to leverage a unified representation combined of the cluster semantics and instances, enhancing the representation with cluster information to facilitate clustering tasks. Formally, for an instance representation $\mathbf{z}_i$, the enhanced representation is given by:
\begin{equation}
    \hat{\mathbf{z}}_i=\left(\mathbf{z}_i+ \operatorname{NN}_{\theta}\left(\mathbf{c}_i \right)\right)/\Vert \mathbf{z}_i+ \operatorname{NN}_{\theta}\left(\mathbf{c}_i \right)\Vert_{2},
\end{equation}
where $\mathbf{c}_i$ represents the cluster assignment of $i$-th instance after Gumbel Softmax. $\operatorname{NN}_{\theta}\left(\cdot \right)$ denotes a single fully connected network, which is the learnable cluster representation. Different from CC which performs contrastive loss on cluster assignment, TCC conducts contrastive loss on the unified representation to better capture cluster semantics. Inspired by TCC, some works~\citep{groupvit2022,ProImp2023} explore the fusion of instance-level and cluster-level representation in various domains. and then conduct contrastive loss on the unified representation, which further explores its effectiveness.

\begin{figure}[t]
   \centering
   \includegraphics[width=0.95\columnwidth]{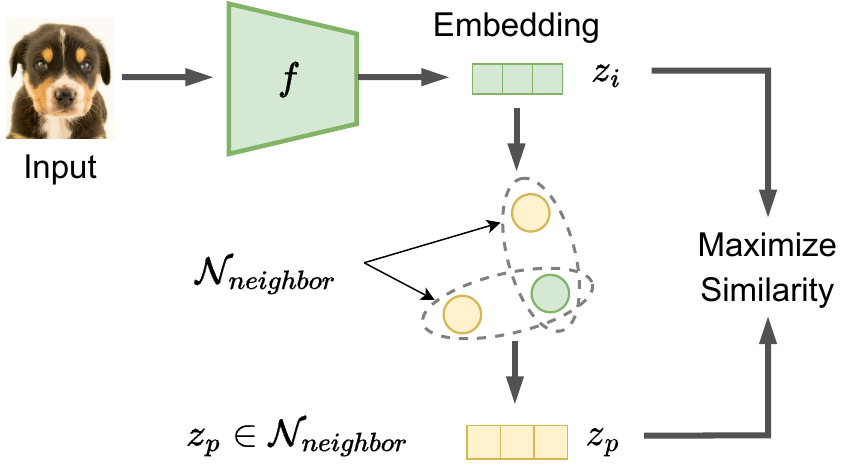}
   \caption{The framework of neighborhood consistency-based methods. Such a paradigm encourages neighboring samples $z_{i}$ and $z_{p}$ in the latent space to have consistent features and cluster assignments, which improves the compactness of clusters.
   }
   \label{fig: Neighborhood_Consistency}
\end{figure}

\subsection{Neighborhood Consistency}
Thanks to the advancements in self-supervised representation learning, the features acquired through discriminative pretext tasks can unveil high-level semantics in the latent space. This provides a crucial prior for clustering, as instances and their neighborhoods in the latent space are likely to belong to the same semantic cluster. Leveraging neighborhood-consistent semantics can further enhance clustering.

SCAN~\citep{SCAN2020} first observes that similar instances will be mapped closely in latent space through self-supervised pretext tasks. Motivated by this observation, SCAN trains a cluster head based on the cluster neighborhood consistency within neighbors. Specifically, SCAN first obtains an encoder $f$ by a pretext task~\citep{PretextRotation2018,PretextTrans2019,PretextInstDist2018,MoCo2020}. It then optimizes a cluster head $h$ by requiring it to make consistent predictions between instances and their nearest neighbors:
\begin{equation}
    \mathcal{L}=-\frac{1}{B}\sum_{i=1}\sum_{j\in\mathcal{N}^k_i}\log\langle \mathbf{p}_{i}, \mathbf{p}_j\rangle - \lambda H(Y).
    \label{eq:scan_loss}
\end{equation}
Here $\mathcal{N}^{k}_{i}$ denotes the $k$-nearest neighbors of the $i$-th instance. The second term in Eq.~\ref{eq:scan_loss} prevents $h$ from assigning all instances to a single cluster which is also used in Eq.~\ref{Eq: cc}.

NNM~\citep{NNM2021} and GCC~\citep{GCC2021} incorporate neighbor information into the framework of contrastive learning to group instances within neighborhoods. In particular, NNM aligns the clustering assignment of an instance with its neighbors through cluster-level contrastive learning:
\begin{equation}
    \mathcal{L}=-\frac{1}{C}\sum_{i=1}^C\log\frac{\exp(\operatorname{s}(\mathbf{q}_i,\mathbf{q}_{\mathcal{N}i}))}{\sum_{j=1}^{C}\exp(\operatorname{s}(\mathbf{q}_i,\mathbf{q}_j))},
    \label{eq:nnm_loss}
\end{equation}
where $\mathbf{q},~\mathbf{q}_{\mathcal{N}}\in\mathbb{R}^{C\times B}$ represent the transpose matrix of $\mathbf{p}$ and $\mathbf{p}_{\mathcal{N}}$, respectively.
In contrast, GCC introduces the graph structure of the latent space to modify the vanilla instance-level contrastive loss. It constructs a normalized symmetric graph Laplacian $\mathbf{L}$ based on the $K$-nn graph:
\begin{equation}
    \begin{aligned}
    \mathbf{L}&=\mathbf{I}-\mathbf{D}^{-\frac{1}{2}}\mathbf{A}\mathbf{D}^{-\frac{1}{2}},\\
    \text{with }
    \mathbf{A}_{ij}&=\begin{cases}
        1, &\text{if } j\in \mathcal{N}^{k}_{i}\text{ or } i\in \mathcal{N}^{k}_{j}\\
        0, &\text{otherwise}
    \end{cases}.
    \end{aligned}    
\end{equation}
Then, the loss function is given by the following form:
\begin{equation}
    \mathcal{L}=-\frac{1}{N}\sum_{i=1}^N\log\frac{\sum_{\mathbf{L}_{ij}<0}-\mathbf{L}_{ij}\exp(\operatorname{s} 
  (\mathbf{z}_i,\mathbf{z}_j)/\tau)}{\sum_{\mathbf{L}_{ij}=0}\exp(\operatorname{s}(\mathbf{z}_i,\mathbf{z}_j)/\tau)},
    \label{eq:gcc_loss}
\end{equation}
where $\tau$ is the temperature. The Graph Laplacian guides the model to attract instances within neighborhoods rather than just augmentation of themselves so that the influence of potential false negative samples~\citep{SURE2022,yang2021MvCLN} can be mitigated. As a result, GCC can better minimize the intra-cluster variance and maximize the inter-cluster variance. The success of this approach has inspired numerous contrastive learning methods~\citep{FNC_Boosting,DIVIDE2024} in various domains to leverage neighbor relationships that effectively address the false negative challenge.

\begin{figure}[t]
   \centering
   \includegraphics[width=\columnwidth]{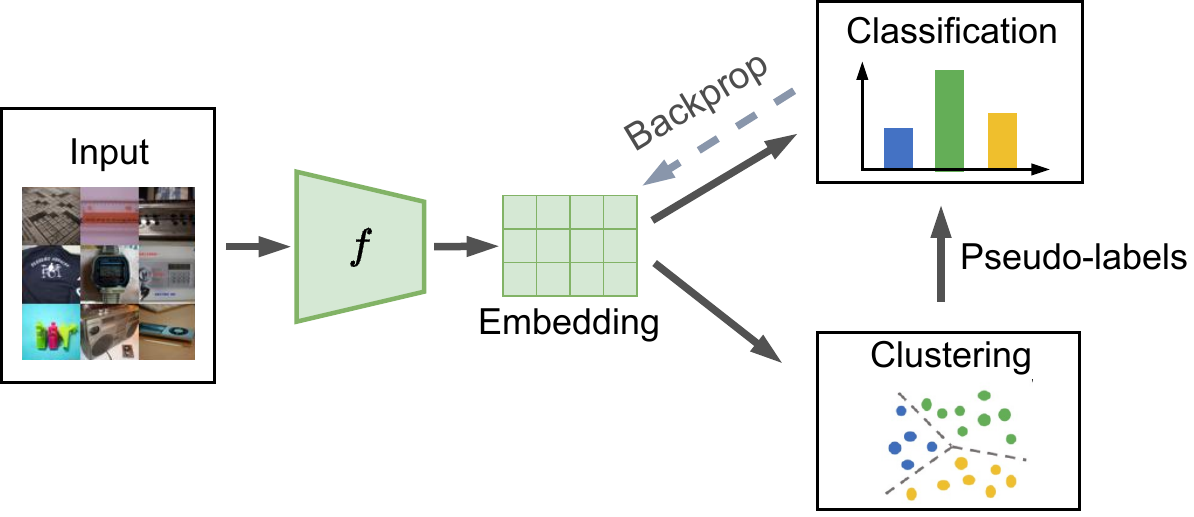}
   \caption{The framework of pseudo-labeling based methods. Given features in the latent space, clustering algorithms such as K-means are performed to get pseudo labels. The pseudo labels, usually filtered by confidence, are then used as supervision signals to guide clustering.}
   \label{fig: Pseudo_Label}
\end{figure}

\subsection{Pseudo-Labeling}
As a prevalent paradigm of semi-supervised classification~\citep{laine2016temporal,MixMatch2019,FixMatch2020}, pseudo-labeling has been extended to deep clustering in recent years. The fundamental assumption of pseudo-labeling is that the predictions on unlabeled data, especially the confident ones, can provide reliable supervision and guide model training. Motivated by this, recent deep clustering works leverage confident predictions to boost clustering performance.

DEC~\citep{DEC2016} is a pioneering work that utilizes labels generated by itself to simultaneously enhance feature representations and optimize clustering assignments. DEC initializes with a pre-trained auto-encoder and $C$ learnable cluster centroids. The soft assignment is calculated using the Student's $t$-distribution, based on the distance between the representation $\mathbf{z}_i$ and centroid $\mathbf{c}_j$:
\begin{equation}
    \mathbf{q}_{ij}=\frac{(1+\|\mathbf{z}_i-\mathbf{c}_j\|^2 / \alpha)^{-\frac{\alpha+1}{2}}}{\sum_{k}(1+\|\mathbf{z}_i-\mathbf{c}_{k}\|^2/ \alpha)^{-\frac{\alpha+1}{2}}},
\end{equation}
where $\alpha$ is the hyper-parameter and $\mathbf{q}_{ij}$ denotes the probability of assigning the instances $i$ to the cluster $j$. DEC refines the clusters by emphasizing the high-confidence assignments and making predictions more confident. Specifically, DEC uses the second power of $\mathbf{q}_i$ as a sharpened assignment to guide the training, \textit{i.e.},
\begin{equation}
\mathbf{p}_{ij}=\frac{\mathbf{q}_{ij}^2 / \operatorname{freq}_j}{\sum_{k}\mathbf{q}_{ik}^2/\operatorname{freq}_{k}},
\end{equation}
 where $\operatorname{freq}_j=\sum_i \mathbf{q}_{ij}$ is the soft cluster frequency and the sharpened assignment is normalized by $f_j$ to prevent feature collapse. 
  Finally, a KL divergence loss between $\mathbf{p}$ and $\mathbf{q}$ minimizes the distances between the two distributions, \textit{i.e.}, $\mathcal{L}=\operatorname{KL}(\mathbf{p}|\mathbf{q})$.

Another notable method of pseudo-labeling is DeepCluster~\citep{DeepCluster2018}. This approach employs K-means clustering on the learned representations to obtain cluster assignments as pseudo-labels. DeepCluster iteratively performs representation learning and clustering in a mutually beneficial manner to bootstrap each other. However, DeepCluster faces limitations in achieving outstanding performance, primarily due to the restricted semantics of the initial representation. {Similar to DeepCluster, ProPos~\citep{ProPos2022} proposes an EM framework of pseudo-labeling, iteratively performing K-means to obtain pseudo labels (E step) and the representation updating (M step). Notably, ProPos significantly outperforms DeepCluster and other methods because ProPos performs K-means on the learned feature of state-of-the-art self-supervised paradigm BYOL~\citep{BYOL2020}. This observation has demonstrated that the semantics of the representation is vital to pseudo-label generation and clustering. Low-quality features would introduce potential noise in pseudo-labels, impact subsequent pseudo-label generation, and mislead representation learning, which accumulates the error in the process.

In addition to the progression of self-supervised paradigms, researchers are actively investigating strategies to alleviate the issue of error accumulation in pseudo-labeling. To be specific,} the challenges in the realm of pseudo-labeling deep clustering remain two-fold: enhancing the accuracy of generating pseudo-labels and maximizing the utility of these pseudo-labels for effective clustering. On the one hand, inaccurate pseudo-labels pose a risk of degradation in clustering performance. On the other hand, determining how to effectively leverage these pseudo-labels for clustering is a critical consideration. These two challenges underscore the ongoing efforts in the pseudo-labeling learning of deep clustering.

The first challenge has been addressed by many works through carefully designing selection methods. For instance, SCAN~\citep{SCAN2020} empirically observed that instances exhibiting highly confident predictions (\textit{i.e.,} $\max(\mathbf{p}_{i})\approx 1$) tend to be correctly clustered by the cluster head. Building on this insight, SCAN opts to choose instances with the most confident predictions as labeled data to fine-tune the model using the cross-entropy loss,
\begin{equation}
    \begin{aligned}
        &\mathcal{L}=\frac{1}{|Y|} \sum_{i\in Y}-\tilde{y}_i\log(\mathbf{p}_i),\\
        &Y=\left\{i\mid\text{conf}_i\geq\eta\right\}, \text{with conf}_i=\max(\mathbf{p}_{i})    
        \label{eq:scan_labeling}
    \end{aligned}
\end{equation}
where $\eta$ is the threshold hyper-parameter to filter the uncertain instances.
TCL~\citep{TCL2022} and SPICE~\citep{SPICE2022} have devised more effective selection strategies to enhance the accuracy of pseudo-labeling. Specifically, TCL selects the most confident predictions as pseudo labels from each cluster $c$:
\begin{equation}
    \begin{aligned}
        &Y^{c}=\left\{\operatorname{topK}(\text{conf}_i)\mid \tilde{y}_i=c\right\}\\
        &Y=\bigcup_{c=1}^{C}Y^{c}
    \end{aligned}
    \label{eq:tcl_labeling}
\end{equation}
where $\operatorname{topK}(\cdot)$ returns the indices of the top $K$ confident instances and $\bigcup$ denotes the union of the pseudo labels from all clusters. Here $K=\gamma N/C$ and $\gamma$ is the selection ratio. The cluster-wise selection leads to more class-balanced pseudo labels compared to threshold-based criteria. It improves the clustering performance, especially for challenging classes.

SPICE introduces a prototype-based pseudo-labeling approach. Specifically, it first re-computes the centroids of each cluster only using the instances with confident predictions, then re-assign each instance with new pseudo labels according to the similarity to the new centroids, formally:
\begin{equation}
\begin{aligned}
    &\mathbf{c}_i^{\prime}=\frac{1}{|Y^{c}|}\sum_{i\in Y^{c}}\mathbf{z}_i,\\
    &\tilde{y}_i^{\prime}=\arg\max_{j}\operatorname{s}(\mathbf{z}_i,\mathbf{c}^{\prime}_{j}).
    \label{eq:spice_labeling}
\end{aligned}
\end{equation}
This operation helps mitigate the influence of potentially incorrect pseudo labels used in calculating centroids, which might accumulate errors in the iterative self-training process.

To address the second challenge, \textit{i.e.}, better utilizing the confident labels, TCL removes negative pairs with the same label in contrastive loss, preventing intra-class instances from pushing apart, \textit{i.e.}, the false negative issue. Meanwhile, SPICE and TCL adopt some semi-supervised classification techniques like FixMatch~\citep{FixMatch2020} that impose the pseudo-label consistency for strong augmentations of the same instance. The marvelous results achieved by these works show the effectiveness of combining reliable pseudo-labeling methods and semi-supervised paradigms in clustering. 

\begin{figure}[t]
   \centering
    \includegraphics[width=\columnwidth]{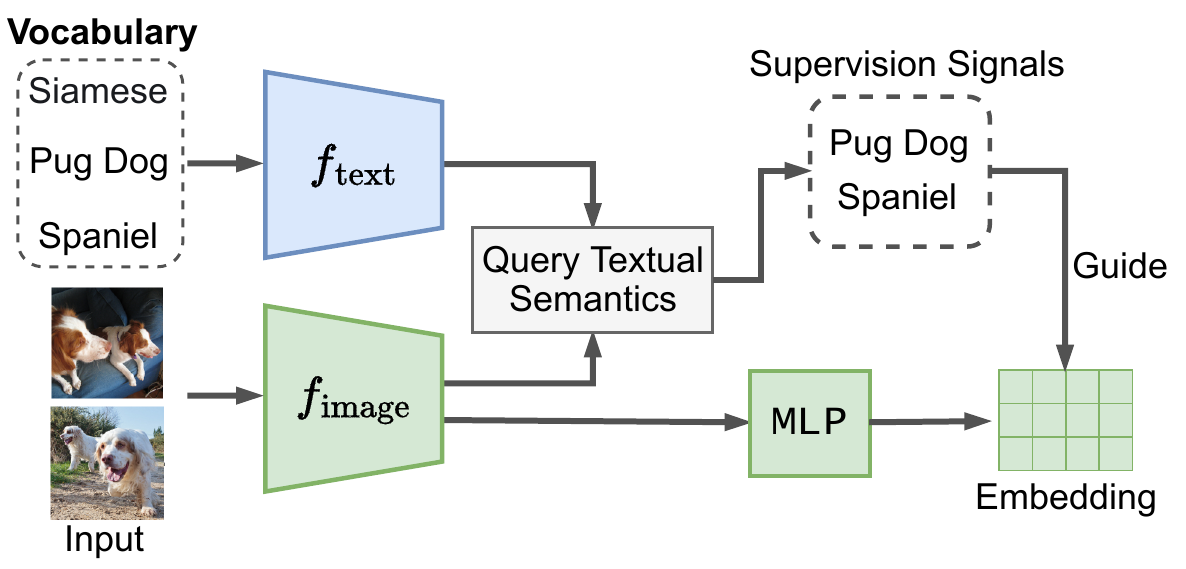}
   \caption{The framework of external knowledge based methods. Instead of mining internal priors from the samples themselves, such a paradigm seeks external information like textual semantics to help distinguish the given samples.
   }
   \label{fig: External_Knowledge}
\end{figure}

\subsection{External Knowledge}
Most clustering approaches focus on grouping data based on inherent characteristics such as structural priors, distribution priors, and augmentation invariance priors. Instead of pursuing internal priors from the data itself, some recent works~\cite{SIC2023, TAC2023} attempt to introduce abundant external knowledge such as textual descriptions to guide clustering. These methods prove effective because the utilization of semantic information from natural language offers valuable supervisory signals that enhance the quality of clustering.

SIC~\citep{SIC2023} is one of the first works in incorporating external knowledge guidance into clustering.
The fundamental concept revolves around generating image pseudo-labels from a textual space pre-trained by CLIP~\citep{radford2021learning}. The process involves three main steps: i) \textit{Construction of Semantic Space}: SIC selects meaningful texts resembling category names to build a semantic space. ii) \textit{Pseudo-labeling}: Pseudo-labels are generated using text semantic centers $\mathbf{h}$ and image representations $\mathbf{z}_i$, formally,
\begin{equation}
\mathbf{q}_i=\text{one-hot}\left(c, \arg\max_l \frac{\exp \left(\mathbf{z}_i^T \mathbf{h}_l\right)}{\Sigma_{l^{\prime}}^c \exp \left(\mathbf{z}_i^T \mathbf{h}_{l^{\prime}}\right)}\right),
\end{equation}
where $c$ is the number of semantic centers, $\mathbf{h}_l$ is the $l$-th center of semantic centers, one-hot operator will generate a $c$-bit one-hot vector. The pseudo-labels is utilized to guide the clustering similar to SCAN~\citep{SCAN2020},
\begin{equation}
\mathcal{L}=\frac{1}{n} \sum_{i=1}^n C E\left(\mathbf{q}_i, \mathbf{p}_i\right),
\end{equation}
where $CE\left(\cdot\right)$ is the cross entropy function.
iii) \textit{Consistency learning}: Enhancing clustering effect by enforcing the consistency between the images and their neighbors in the image space,
\begin{equation}
\mathcal{L}=-\frac{1}{n} \sum_{i=1}^n \log \mathbf{p}_i^T \mathbf{p}_j,
\end{equation}
where $j$ is an instance index randomly selected from the nearest neighbors $\mathcal{N}_k\left(\mathbf{z}_i\right)$ of $i$-th instance. Note that, SIC essentially pulls image embeddings closer to embeddings in semantic space, while ignoring the improvement of text semantic embeddings.

\begin{table*}[t]
\caption{The summary of deep clustering methods from the perspective of prior knowledge.}
\label{tab:summary}
\tablestyle{7pt}{2.5}{ 

\begin{tabular}{|c|l|l|l|}
\hline
\multicolumn{2}{|c|}{Prior Knowledge}  & Method & Major Contribution    \\ \hline
 \multirow{5}{*}{\makecell*[c]{Structurture\\ Prior}} &
  % The latent feature is well-structured &
  \multirow{5}{*}{\makecell*[l]{Inherent data structure \\reflect semantic relation}} &
  \makecell[l]{ABDC (\citeyear{ABDC2013})} &
  \makecell[l]{optimize features and clustering assignment in an\\ EM manner} \\ \cline{3-4} 
 &
   &
  \makecell[l]{DEN (\citeyear{DEN2014}),\\ SpectralNet (\citeyear{SpectralNet2018})} &
  \makecell[l]{extend spectral clustering from shallow to deep} \\ \cline{3-4} 
 &
   &
  \makecell[l]{PARTY (\citeyear{PARTY2016})} &
  \makecell[l]{introduce the sparsity prior from subspace learning \\to deep clustering} \\ \cline{3-4} 
 &
   &
   \makecell[l]{JULE (\citeyear{JULE2016})} &
  \makecell[l]{extend agglomerative clustering from shallow to deep} \\
 \cline{3-4} 
 &
   &
  \makecell[l]{DCC (\citeyear{DCC2018})} &
  \makecell[l]{propose relation matching to achieve non-parametric \\deep clustering} \\ \hline
\multirow{2}{*}{\makecell*[c]{Distribution\\ Prior}} &
  \multirow{2}{*}{\makecell*[l]{Instances of different \\semantics follow distinct \\data distribution}} &
  \makecell[l]{VaDE (\citeyear{VaDE2016})} &
  \makecell[l]{learn distinct cluster distributions by Gaussian \\mixture model} \\ \cline{3-4} 
 &
   &
  \makecell[l]{ClusterGAN (\citeyear{ClusterGAN2019}) \\ DCGAN (\citeyear{DCGAN2016})} &
  \makecell[l]{implicitly learn cluster distribution with GAN} \\ \hline
\multirow{5}{*}{\makecell*[c]{Augmentation\\ Invariance}} &
  \multirow{2}{*}{\makecell*[l]{Instance features are \\invariant to data \\augmentation}} &
  \makecell[l]{IMSAT (\citeyear{IMSAT2017})} &
  \makecell[l]{propose the invariance between pair-wise augmented \\samples} \\ \cline{3-4} 
 &
   &
  \makecell[l]{IIC (\citeyear{IIC2019}),\\Completer (\citeyear{Completer2021})} &
  \makecell[l]{propose the mutual information framework with \\respect to augmentation invariance} \\ \cline{2-4} 
 & \multirow{3}{*}{\makecell*[l]{Cluster assignments \\ are invariant to data \\augmentation}}
   &
  \makecell[l]{PICA (\citeyear{PICA2020})} &
  \makecell[l]{explore invariance between cluster assignments of \\augmented samples} \\ \cline{3-4} 
 &
   &
  \makecell[l]{CC (\citeyear{CC2021}),\\ DRC (\citeyear{DRC2020})} &
  \makecell[l]{simultaneously explore augmentation invariance at \\instance and cluster level} \\ \cline{3-4} 
 &
   &
  \makecell[l]{TCC (\citeyear{TCC2021})} &
  \makecell[l]{leverage a unified representation combined of the\\cluster semantics and instances} \\ \hline
\multirow{3}{*}{\makecell[c]{Neighborhood\\ Consistency}} &
  \multirow{3}{*}{\makecell[l]{Neighboring instances \\have similar semantics}} &
  \makecell[l]{SCAN (\citeyear{SCAN2020})} &
  \makecell[l]{impose consistent cluster assignments between \\neighboring instances} \\ \cline{3-4} 
 &
   &
  \makecell[l]{NNM (\citeyear{NNM2021})} &
  \makecell[l]{perform cluster-level contrastive learning between \\neighbors} \\ \cline{3-4} 
 &
   &
  \makecell[l]{GCC (\citeyear{GCC2021})} &
  \makecell[l]{perform instance- and cluster-level contrastive \\learning between neighbors} \\ \hline
\multirow{6}{*}{\makecell*[c]{Pseudo Label}} &
  \multirow{6}{*}{\makecell*[l]{Cluster assignments \\with high confidence \\are reliable}} & 
  \makecell[l]{DEC (\citeyear{DEC2016})} &
  \makecell[l]{construct target cluster distribution via sharpening} \\ \cline{3-4} 
 &
   &
  \makecell[l]{DeepCluster (\citeyear{DeepCluster2018})} &
  \makecell[l]{generate pseudo labels with K-means} \\ \cline{3-4} 
 &
   &
  \makecell[l]{SCAN (\citeyear{SCAN2020})} &
  \makecell[l]{select high-confident predictions and finetune the \\model with strong augmented samples} \\ \cline{3-4} 
 &
   &
  \makecell[l]{SPICE (\citeyear{SPICE2022})} &
  \makecell[l]{select pseudo labels with the help of prototypes and \\adopt semi-supervised learning to fine-tune the model} \\ \cline{3-4} 
 &
   &
  \makecell[l]{TCL (\citeyear{TCL2022})} &
  \makecell[l]{use pseudo labels to mitigate false negative pairs in \\contrastive learning} \\ \cline{3-4} 
 &
   &
  \makecell[l]{ProPos (\citeyear{ProPos2022})} &
  \makecell[l]{use pseudo label from K-means to increase cluster \\compactness} \\ \hline
  \multirow{2}{*}{\makecell[c]{External\\ Knowledge}}      & \multirow{2}{*}{\makecell[l]{Abundant clustering-\\favorable knowledge \\exists in open world}}    & \makecell[l]{SIC (\citeyear{SIC2023})}    &  \makecell[l]{generate image pseudo labels from the textual space\\ from pre-trained vision-language model}  \\\cline{3-4} 
     &       & \makecell[l]{TAC (\citeyear{TAC2023})}    &  \makecell[l]{construct more discriminative text counterparts and \\perform cross-modal distillation to improve clustering}       \\ \hline
\end{tabular}}
\end{table*}

TAC~\citep{TAC2023} focuses on leveraging textual semantics to enhance the feature discriminability. Specifically, it retrieves a text counterpart among representative nouns for each image, which improves K-means performance without any additional training. Besides, TAC proposes a mutual distillation paradigm to incorporate the image and text modalities, which further improves the clustering performance. The cross-modal mutual distillation strategy is formulated as follows:
\begin{equation}
\begin{aligned}
    \mathcal{L}&=\sum_{i=1}^C \mathcal{L}_i^{v\rightarrow t}+\mathcal{L}_i^{t\rightarrow v},\\
 L_i^{v \rightarrow t}&=-\log \frac{\exp \left(\operatorname{sim}\left(\hat{\mathbf{q}}_i, \hat{\mathbf{p}}_i^{\mathcal{N}}\right) / \tau\right)}{\sum_{k=1}^K \exp \left(\operatorname{sim}\left(\hat{\mathbf{q}}_i, \hat{\mathbf{p}}_k^{\mathcal{N}}\right) / \tau\right)}, \\
 L_i^{t \rightarrow v}&=-\log \frac{\exp \left(\operatorname{sim}\left(\hat{\mathbf{p}}_i, \hat{\mathbf{q}}_i^{\mathcal{N}}\right) / \hat{\tau}\right)}{\sum_{k=1}^K \exp \left(\operatorname{sim}\left(\hat{\mathbf{p}}_i, \hat{\mathbf{q}}_k^{\mathcal{N}}\right) / \tau\right)},
\end{aligned}
\end{equation}
where $\tau$ is the softmax temperature parameter, $\hat{\mathbf{p}}_i,\hat{\mathbf{q}}_i\in \mathbb{R}^{1 \times N}$ is the $i$-th column of image and text assignment matrix, $\hat{\mathbf{p}}_i^{\mathcal{N}}, \hat{\mathbf{q}}_i^{\mathcal{N}}\in \mathbb{R}^{1 \times N}$ is the $i$-th column of image and text random nearest neighbor matrix. The mutual distillation strategy has two advantages. On the one hand, it generates more discriminative clusters through cluster-level contrastive loss. On the other hand, it encourages consistent clustering assignments between each sample and its cross-modal neighbors, which bootstraps the clustering performance in both modalities.

\newpage
\section{Experiment}
\label{sec:expmt}
In this section, we introduce the evaluation of deep clustering. Briefly, we first present the evaluation metrics and common benchmarks. Then we analyze the results of the existing deep clustering methods.

\subsection{Evaluation Metrics}
For clustering evaluation, three metrics are commonly used to measure how the predicted cluster assignments $\tilde{y}$ match the ground truth labels $y$, including accuracy (ACC), normalized mutual information (NMI), and adjusted rand index (ARI). A higher value of the metrics corresponds to better clustering performance. The definitions of the three metrics are as follows:
\begin{itemize}
    \item ACC~\citep{ACC2009} indicates the correct rate of clustering predictions:
    \begin{equation}
        \operatorname{ACC}=\frac{1}{N}\sum_{i=1}^N \mathbf{1}\{y_i=\tilde{y}_i\},
        \label{Eq: ACC}
    \end{equation}
    where the Hungarian matching~\citep{Hungarian1955} is first applied to align the predictions and labels.
    \item NMI~\citep{NMI2011} quantifies the mutual information between the predicted labels $\tilde{\mathbf{Y}}$ and ground truth labels $\mathbf{Y}$:
    \begin{equation}
        \operatorname{NMI}=\frac{I(\tilde{\mathbf{Y}}; \mathbf{Y})}{\frac{1}{2}[H(\tilde{\mathbf{Y}})+H(\mathbf{Y})]},
        \label{Eq: NMI}
    \end{equation}
    where $H(\mathbf{Y})$ denotes the entropy of $Y$ and $I(\tilde{\mathbf{Y}}; \mathbf{Y})$ denotes the mutual information between $\tilde{\mathbf{Y}}$ and $\mathbf{Y}$.
    \item ARI~\citep{ARI1985} is the normalization of the rand index (RI), which counts the number of instances pairs in the same cluster and different clusters:
    \begin{equation}
        \operatorname{RI}=\frac{\operatorname{TP}+{\operatorname{TN}}}{\operatorname{C}^2_N},
        \label{Eq: RI}
    \end{equation}
    where $\operatorname{TP}$ and $\operatorname{TN}$ refer to the number of true positive pairs and true negative pairs, $\operatorname{C}^2_N$ is the number of possible instance pairs. ARI is computed by adding the following normalization:
    \begin{equation}
        \operatorname{ARI}=\frac{\operatorname{RI}-\mathbb{E}(\operatorname{RI})}{\operatorname{max}(\operatorname{RI})-\mathbb{E}\left(\operatorname{RI}\right)},
        \label{Eq: ARI}
    \end{equation}
    where $\mathbb{E}(\operatorname{RI})$ denotes the expectation of RI.
\end{itemize}

\begin{table}[t]
\caption{A summary of datasets commonly used for deep clustering.}
\centering
\tablestyle{4pt}{1.5}{
\begin{tabular}{@{}ccccc@{}}
\toprule
Dataset        & Split       & Samples   & Classes & Image Size     \\ \midrule
CIFAR-10       & Train+Test  & 60,000    & 10      & 32$\times$32   \\
CIFAR-100      & Train+Test  & 60,000    & 20      & 32$\times$32   \\
STL-10         & Train+Test  & 13,000    & 10      & 96$\times$96   \\
ImageNet-10    & Train       & 13,000    & 10      & 96$\times$96   \\
ImageNet-Dogs  & Train       & 19,500    & 15      & 96$\times$96   \\ \midrule
Tiny-ImageNet  & Train       & 100,000   & 200     & 64$\times$64   \\
ImageNet-1K    & Train       & 1,281,167 & 1000    & 224$\times$224 \\\bottomrule
\end{tabular}%
}
\label{tab:datasets}
\end{table}

\begin{table*}[t]
\caption{Clustering performance on five widely-used image clustering datasets. SCAN$^*$ denotes the clustering results using only neighborhood consistency loss without the self-labeling step. $\dagger$ denotes using the train and test split for training and testing respectively, instead of using both splits for training and testing. Horizontal lines separate methods with different priors. From top to bottom are structure prior, distribution prior, augmentation invariance, neighborhood consistency, pseudo-labeling, and external knowledge.}
\label{tab: cluster_performance}
\centering
\tablestyle{4pt}{1.8}{
\begin{tabular}{c|ccc|ccc|ccc|ccc|ccc}
\toprule
\multirow{2}{*}{Method} &
  \multicolumn{3}{c|}{CIFAR-10} &
  \multicolumn{3}{c|}{CIFAR-100} &
  \multicolumn{3}{c|}{STL-10} &
  \multicolumn{3}{c|}{ImageNet-10} &
  \multicolumn{3}{c}{ImageNet-Dogs}  \\
               & ACC  & NMI  & ARI  & ACC  & NMI  & ARI  & ACC  & NMI  & ARI  & ACC  & NMI  & ARI  & ACC  & NMI  & ARI \\ \hline
K-means (\citeyear{kmeans1967})      & 22.9 & 8.7  & 4.9  & 13.0 & 8.4  & 2.8  & 19.2 & 12.5 & 6.1  & 24.1 & 11.9 & 5.7  & -    & -    & -  \\ \hline
JULE (\citeyear{JULE2016})           & 27.2 & 19.2 & 13.8 & 13.7 & 10.3 & 3.3  & 27.7 & 18.2 & 16.4 & 30.0 & 17.5 & 13.8 & 13.8 & 5.4  & 2.8  \\ \hline
DCGAN (\citeyear{DCGAN2016})          & 31.5 & 26.5 & 17.6 & 15.1 & 12.0 & 4.5  & 29.9 & 22.7 & 16.2 & 31.3 & 18.6 & 14.2 & 17.8 & 9.8  & 7.3  \\ \hline
IIC (\citeyear{IIC2019})            & 61.7 & 51.3 & 41.1 & 25.7 & 22.5 & 11.7 & 59.6 & 49.6 & 39.7 & -    & -    & -    & -    & -    & -   \\
PICA (\citeyear{PICA2020})           & 69.6 & 59.1 & 51.2 & 33.7 & 31.0 & 17.1 & 71.3 & 61.1 & 53.1 & 87.0 & 80.2 & 76.1 & 35.3 & 35.2 & 20.1  \\ 
CC (\citeyear{CC2021})            & 79.0 & 70.5 & 63.7 & 42.9 & 43.1 & 26.6 & 85.0 & 76.4 & 72.6 & 89.3 & 85.9 & 82.2 & 42.9 & 44.5 & 27.4  \\
TCC (\citeyear{TCC2021})            & 90.6 & 79.0 & 73.3 & 49.1 & 47.9 & 31.2 & 81.4 & 73.2 & 68.9 & 89.7 & 84.8 & 82.5 & 59.5 & 55.4 & 41.7   \\ \hline
SCAN$^*$ (\citeyear{SCAN2020})       & 81.8 & 71.2 & 66.5 & 42.2 & 44.1 & 26.7 & 75.5 & 65.4 & 59.0 & -    & -    & -    & -    & -    & -      \\
NNM$^\dagger$ (\citeyear{NNM2021})  & 83.7 & 73.7 & 69.4 & 45.9 & 48.0 & 30.2 & 76.8 & 66.3 & 59.6 & -    & -    & -    & 58.6 & 60.4 & 44.9   \\
GCC (\citeyear{GCC2021})            & 85.6 & 76.4 & 72.8 & 47.2 & 47.2 & 30.5 & 78.8 & 68.4 & 63.1 & 90.1 & 84.2 & 82.2 & 52.6 & 49.0 & 36.2   \\ \hline
DEC (\citeyear{DEC2016})            & 30.1 & 25.7 & 16.1 & 18.5 & 13.6 & 5.0  & 35.9 & 27.6 & 18.6 & 38.1 & 28.2 & 20.3 & 19.5 & 12.2 & 7.9 \\
DeepCluster (\citeyear{DeepCluster2018})    & 37.4 & -    & -    & -    & -    & -    & 33.4 & -    & -    & 18.9 & -    & -    & -    & -    & -     \\
SCAN$^\dagger$ (\citeyear{SCAN2020}) & 87.6 & 78.7 & 75.8 & 48.3 & 48.5 & 31.4 & 81.8 & 70.3 & 66.1 & -    & -    & -    & 59.3 & 61.2 & 45.7   \\
SPICE (\citeyear{SPICE2022})         & 83.8 & 73.4 & 70.5 & 46.8 & 44.8 & 29.4 & 90.8 & 81.7 & 81.2 & 92.1 & 82.8 & 83.6 & 64.6 & 57.2 & 47.9   \\
TCL (\citeyear{TCL2022})            & 88.7 & 81.9 & 78.0 & 53.1 & 52.9 & 35.7 & 86.8 & 79.9 & 75.7 & 89.5 & 87.5 & 83.7 & 64.4 & 62.3 & 51.6   \\
ProPos (\citeyear{ProPos2022})         & 94.3 & 88.6 & 88.4 & 61.4 & 60.6 & 45.1 & 86.7 & 75.8 & 73.7 & 95.6 & 89.6 & 90.6 & 74.5 & 69.2 & 62.7 \\ \hline
SIC$^\dagger$ (\citeyear{SIC2023})  & 92.6 & 84.7 & 84.4 & 58.3 & 59.3 & 43.9 & 98.1 & 95.3 & 95.9 & 98.2 & 97.0 & 96.1 & 69.7 & 69.0 & 55.8   \\
TAC (\citeyear{TAC2023})            & 92.3 & 84.1 & 83.9 & 60.7 & 61.1 & 44.8 & 98.2 & 95.5 & 96.1 & 99.2 & 98.5 & 98.3 & 83.0 & 80.6 & 72.2   \\ \bottomrule
\end{tabular}%
}
\end{table*}
\subsection{Datasets}
In the early stage, deep clustering methods are evaluated on relatively small and low-dimensional datasets (\textit{e.g.} COIL-20~\citep{COIL20_1996}, YaleB~\citep{YaleB2001}). Recently, with the rapid development of deep clustering methods, it has become more popular to evaluate clustering performance on more complex and challenging datasets. There are five widely used benchmark datasets:
\begin{itemize}
    \item CIFAR-10~\citep{CIFAR2009} consists of 60,000 colored images from 10 different classes including airplane, automobile, bird, cat, deer, dog, frog, horse, ship, and truck.

    \item CIFAR-100~\citep{CIFAR2009} contains 100 classes grouped into 20 superclasses. Each image comes with a ``fine'' class label and a ``coarse'' superclass label.

    \item STL-10~\citep{STL2011} contains 13,000 labeled images from 10 object classes. Besides, it provides 100,000 unlabeled images for self-supervised learning to enhance the clustering performance.

    \item ImageNet-10~\citep{ImageNetSubset2017} is a subset of the ImageNet dataset~\citep{ImageNet2009}. It contains 10 classes, each with 1,300 high-resolution images.

    \item ImageNet-Dog~\citep{ImageNetSubset2017} is another subset of ImageNet. It consists of images belonging to 15 dog breeds, which is suitable for fine-grained clustering tasks.
\end{itemize}

Apart from them, some recent works employ two more challenging large-scale datasets, Tiny-ImageNet~\citep{TinyImageNet2015} and ImageNet-1K~\citep{ImageNet2009}, to evaluate the effectiveness and efficiency. A brief description of these datasets is summarized in Table~\ref{tab:datasets}.

\subsection{Performance Comparisons}
The clustering performance on five widely used datasets is shown in Table~\ref{tab: cluster_performance}. Thanks to the feature extraction ability of deep neural networks, early deep clustering methods based on structure and distribution priors achieve much better performance than the classic K-means. Then, a series of contrastive clustering methods significantly improve the performance by introducing additional priors through data augmentation. After that, more advanced methods boost the performance by further considering the neighborhood consistency (GCC compared with CC) and utilizing pseudo labels (SCAN compared with SCAN$^*$). Notably, the performance gains of different priors are independent. For example, ProPos remarkably outperforms DEC and CC by additionally utilizing the augmentation invariance or pseudo-labeling priors, respectively. Very recently, external-knowledge-based methods achieved state-of-the-art performance, which proves the promising prospect of such a new deep clustering paradigm. In addition, clustering becomes more challenging when the category number grows (from CIFAR-10 to CIFAR-100) or the semantics becomes more complex (from CIFAR-10 to ImageNet-Dogs). Such results indicate that more challenging datasets such as full ImageNet-1K are expected to benchmark in future works.

\section{Application in Vicinagearth}
\label{sec:appli}
In this section, we explore some typical applications of deep clustering within the domain of Vicinagearth, a term crafted from the fusion of "Vicinage" and "Earth." 
Vicinagearth represents the critical spatial expanse ranging from 1,000 meters below sea level (the depth at which sunlight ceases to penetrate) to 10,000 meters above sea level (the typical cruising altitude of commercial aircraft).
This zone is of great importance as it encompasses the core regions of human activity including areas of habitation and production. 
Recently, deep clustering has emerged as an indispensable analytical tool within Vicinagearth, instrumental in unveiling complex patterns and structures of data within the vicinal space. The diverse applications of deep clustering in this zone include anomaly detection, environmental monitoring, community detection, person re-identification, and more.

\textbf{Anomaly Detection},
% Anomaly Detection (a.k.a. Outlier Detection, Novelty Detection) is a technique for identifying abnormal instances or patterns among data. Deep clustering can be employed to analyze sensor data from various sources \citep{ano_det2022}, such as underwater monitoring systems, aerial sensors, or ground-based sensors. By identifying patterns and learning the normal behavior of the environment, the system can detect anomalies indicative of security threats or irregular activities.
also known as Outlier Detection~\citep{comaniciu2002mean} or Novelty Detection~\citep{ester1996density}, attempts to identify abnormal instances or patterns. In the context of Vicinagearth, deep clustering proves valuable for analyzing sensor data obtained from diverse sources like underwater monitoring systems, aerial sensors, or ground-based sensors~\citep{ano_det2022}. Through the analysis of the patterns and typical behaviors from the sensor data, the system becomes adept at detecting anomalies, which may signal security threats or irregular activities.

\textbf{Environmental Monitoring}
involves the analysis of data collected from environmental sensors~\citep{xia2007near}, such as monitoring air quality, water conditions, and geological factors. The primary goal is to ensure the health of ecosystems~\citep{wu2016data} and detect potential environmental threats, such as pollution events or natural disasters. Deep clustering techniques play a crucial role in grouping similar environmental patterns, facilitating the identification of abnormalities. This application contributes to real-time environmental monitoring~\citep{kumar2012environmental}, enhancing the ability to respond promptly to environmental challenges.

\textbf{Community Detection} ~\citep{fortunato2010community,jin2021survey}
involves evaluating how groups of nodes are clustered or partitioned and their tendency to strengthen or break apart within a network. In the context of Vicinagearth, this technique is applied to identify groups of species~\citep{murdock2011identifying} that interact closely or share similar ecological niches. Deep clustering plays a pivotal role in the analysis of complex ecological networks~\citep{montoya2006ecological}, contributing to a deeper understanding of ecological communities and their dynamics.

\textbf{Person Re-identification}~\citep{wu2019deep,reid2021}
is a crucial task that involves recognizing and matching individuals across different camera views~\citep{yang2022DART}. This technology plays a significant role in public safety and law enforcement initiatives, as it helps to monitor densely populated areas for including potential threats or subjects on the watchlist. The integration of deep clustering algorithms has remarkedly improved the scalability and efficiency~\citep{yan2023efficient} of person re-identification systems. Deep clustering effectively enables the management of the complexities presented by large and dynamically changing crowds.  Furthermore, the adaptability of deep clustering techniques broadens their use to include the monitoring of natural habitats and the tracking of wildlife in diverse and uncontrolled settings.

\section{Future Challenges}
\label{sec:chall}
Although existing works achieve remarkable performance, some practical challenges and emerging requirements have yet to be fully addressed. In this section, we delve into some future directions of modern deep clustering.

\subsection{Fine-grained Clustering}
The objective of fine-grained clustering is to discern subtle and intricate variations within data, which is particularly advantageous in research like the identification of biological subspecies~\citep{ScBatchEff2023,SCbridge2023}. The primary challenge is that fine-grained classes exhibit a high degree of similarity, where distinctions often lie in coloration, markings, shape, or other subtle characteristics. In such scenarios, traditional coarse-grained clustering priors frequently prove inadequate. For instance, color and shape augmentations in augmentation invariance prior would become ineffective. Recently, C3-GAN~\citep{C3GAN2022} employs contrastive learning within adversarial training to generate lifelike images, enabling the nuanced capture of fine-grained details and ensuring the separability between clusters.

\subsection{Non-parametric Clustering}
Many clustering methods typically require a predefined and fixed number of clusters. However, real-world datasets often present a challenge with an unknown number of clusters, reflecting situations closer to reality. 
Only a few works~\citep{chen2015deep,DCC2018, zhao2019streaming,wang2021dnb} have been devoted to solving this problem. These methods often rely on calculating global similarity and introduce huge computational costs, especially in large-scale datasets. Therefore, efficiently determining the optimal value of cluster number $C$ remains an open challenge, often involving the incorporation of human priors. Among existing works, DeepDPM introduces Dirichlet Process Gaussian Mixture Models (DPGMM)~\citep{antoniak1974mixtures} that utilize the Dirichlet Process as the prior distribution over mixture components. DeepDPM dynamically adjusts the number of clusters $C$ through split and merge operations guided by the Metropolis-Hastings framework~\citep{hastings1970monte}. 

\subsection{Fair Clustering}
Collecting Real-world datasets from diverse sources with various acquisition methods can enhance the generalization of machine learning models. However, these datasets frequently manifest inherent biases, notably in sensitive attributes such as gender, race, and ethnicity. These biases would introduce disparities among individuals and minority groups, leading to cluster partitions that deviate from the underlying objective characteristics of the data. The pursuit of fairness is particularly pertinent in applications where unbiased and equitable analyses are crucial, such as employment, healthcare, and education. To tackle this challenge, fair clustering seeks to mitigate the influence of these biases given the biased attributes for each sample.

To address this daunting task,~\citeauthor{chierichetti2017fair} first introduces a data pre-processing method known as fairlet decomposition. Recent advancements address this issue on large-scale data through adversarial training~\citep{li2020deep} and mutual information maximization~\citep{zeng2023deep}. Notably, \citeauthor{zeng2023deep} designs a novel metric that assesses both clustering quality and fairness from the perspective of information theory. Despite these developments, there is still room for improvement, and the establishment of better evaluation metrics is a continuing area of this research.

\subsection{Multi-view Clustering}
Multi-view data~\citep{xu2013,liu2019multiple} is common in real-world situations where information is captured from a variety of sensors or observed from multiple angles. This data is inherently rich, offering diverse yet consistent information. For example, an RGB view would provide color details while the depth view reveals spatial information, which represents the complementary aspects of the views. Simultaneously, there exists a level of view consistency as the same object possesses common attributes across different views.
To deal with multi-view data, multi-view clustering~\citep{deng2015,liutong2019} is proposed to exploit both the complementary and consistent characters. The goal is to integrate information from all views to produce a unified and insightful clustering result. 

Over recent years, several deep-learning approaches~\citep{DCCA2013,DCCAE2016,MKK, COMIC2019} have been developed to address this challenge. Binary multi-view clustering~\citet{BMVC2018} simultaneously refines binary cluster structures alongside discrete data representations, ensuring cohesive clustering. In pursuit of view consistency,~\citet{Completer2021,lin2022dcp} maximize mutual information across views, thus aligning common properties. SURE~\citep{SURE2022} aims to strengthen the consistency of shared features between views by utilizing robust contrastive loss. Recently,~\citet{ProImp2023} performs bound contrastive loss to preserve the view complementary at the cluster level. These innovative methodologies demonstrate the significant strides made in the field of multi-view analysis, where clustering continues to play a pivotal role in enhancing the synergistic exploitation of multi-view data.

\section{Conclusion}
The key to deep clustering or unsupervised learning is to seek effective supervision to guide representation learning. Different from traditional taxonomies from the network structure or data type, this survey offers a comprehensive review from the perspective of prior knowledge. With the evolution of clustering technologies, there is a discernible trend shifting from exploring priors within the data itself to external knowledge like natural language guiding. The exploration of external pre-trained models like ChatGPT or GPT-4V(ision) might emerge as a promising avenue. This survey potentially provides some valuable insight and inspires further exploration and advancements in deep clustering.
% \begin{appendices}

% \end{appendices}

\bibliography{sn-bibliography}% common bib file

\begin{thebibliography}{125}
\providecommand{\natexlab}[1]{#1}
\providecommand{\url}[1]{{#1}}
\providecommand{\urlprefix}{URL }
\providecommand{\doi}[1]{\url{https://doi.org/#1}}
\providecommand{\eprint}[2][]{\url{#2}}
 \bibcommenthead

\bibitem[{Amig{\'o} et~al(2009)Amig{\'o}, Gonzalo, Artiles, and Verdejo}]{ACC2009}
Amig{\'o} E, Gonzalo J, Artiles J, et~al (2009) A comparison of extrinsic clustering evaluation metrics based on formal constraints. Information retrieval 12:461--486

\bibitem[{Andrew et~al(2013)Andrew, Arora, Bilmes, and Livescu}]{DCCA2013}
Andrew G, Arora R, Bilmes J, et~al (2013) Deep canonical correlation analysis. In: International conference on machine learning, PMLR, pp 1247--1255

\bibitem[{Antoniak(1974)}]{antoniak1974mixtures}
Antoniak CE (1974) Mixtures of dirichlet processes with applications to bayesian nonparametric problems. The annals of statistics pp 1152--1174

\bibitem[{Belkin and Niyogi(2001)}]{LEcluster2001}
Belkin M, Niyogi P (2001) Laplacian eigenmaps and spectral techniques for embedding and clustering. Advances in neural information processing systems 14

\bibitem[{Bengio et~al(2013)Bengio, Courville, and Vincent}]{AE2013}
Bengio Y, Courville A, Vincent P (2013) Representation learning: A review and new perspectives. IEEE transactions on pattern analysis and machine intelligence 35(8):1798--1828

\bibitem[{Berthelot et~al(2019)Berthelot, Carlini, Goodfellow, Papernot, Oliver, and Raffel}]{MixMatch2019}
Berthelot D, Carlini N, Goodfellow I, et~al (2019) Mixmatch: A holistic approach to semi-supervised learning. Advances in neural information processing systems 32

\bibitem[{Cai et~al(2023)Cai, Qiu, Chen, Zhang, and Chen}]{SIC2023}
Cai S, Qiu L, Chen X, et~al (2023) Semantic-enhanced image clustering. In: Proceedings of the AAAI Conference on Artificial Intelligence, pp 6869--6878

\bibitem[{Caron et~al(2018)Caron, Bojanowski, Joulin, and Douze}]{DeepCluster2018}
Caron M, Bojanowski P, Joulin A, et~al (2018) Deep clustering for unsupervised learning of visual features. In: Proceedings of the European conference on computer vision (ECCV), pp 132--149

\bibitem[{Chang et~al(2017)Chang, Wang, Meng, Xiang, and Pan}]{ImageNetSubset2017}
Chang J, Wang L, Meng G, et~al (2017) Deep adaptive image clustering. In: Proceedings of the IEEE international conference on computer vision, pp 5879--5887

\bibitem[{Chatterjee and Ahmed(2022)}]{ano_det2022}
Chatterjee A, Ahmed BS (2022) Iot anomaly detection methods and applications: A survey. Internet of Things 19:100568

\bibitem[{Chen(2015)}]{chen2015deep}
Chen G (2015) Deep learning with nonparametric clustering. arXiv preprint arXiv:150103084

\bibitem[{Chierichetti et~al(2017)Chierichetti, Kumar, Lattanzi, and Vassilvitskii}]{chierichetti2017fair}
Chierichetti F, Kumar R, Lattanzi S, et~al (2017) Fair clustering through fairlets. Advances in neural information processing systems 30

\bibitem[{Coates et~al(2011)Coates, Ng, and Lee}]{STL2011}
Coates A, Ng A, Lee H (2011) An analysis of single-layer networks in unsupervised feature learning. In: Proceedings of the fourteenth international conference on artificial intelligence and statistics, JMLR Workshop and Conference Proceedings, pp 215--223

\bibitem[{Comaniciu and Meer(2002)}]{comaniciu2002mean}
Comaniciu D, Meer P (2002) Mean shift: A robust approach toward feature space analysis. IEEE Transactions on pattern analysis and machine intelligence 24(5):603--619

\bibitem[{Dang et~al(2021)Dang, Deng, Yang, Wei, and Huang}]{NNM2021}
Dang Z, Deng C, Yang X, et~al (2021) Nearest neighbor matching for deep clustering. In: Proceedings of the IEEE/CVF Conference on Computer Vision and Pattern Recognition, pp 13693--13702

\bibitem[{Deng et~al(2015)Deng, Lv, Liu, Huang, Tao, and Gao}]{deng2015}
Deng C, Lv Z, Liu W, et~al (2015) Multi-view matrix decomposition: A new scheme for exploring discriminative information. In: Twenty-Fourth International Joint Conference on Artificial Intelligence

\bibitem[{Deng et~al(2009)Deng, Dong, Socher, Li, Li, and Fei-Fei}]{ImageNet2009}
Deng J, Dong W, Socher R, et~al (2009) Imagenet: A large-scale hierarchical image database. In: 2009 IEEE conference on computer vision and pattern recognition, Ieee, pp 248--255

\bibitem[{Dong et~al(2021)Dong, Wang, and Abbas}]{SurveyOnApp2022}
Dong S, Wang P, Abbas K (2021) A survey on deep learning and its applications. Computer Science Review 40:100379

\bibitem[{Ester et~al(1996{\natexlab{a}})Ester, Kriegel, Sander, Xu et~al}]{DBSCAN1996}
Ester M, Kriegel HP, Sander J, et~al (1996{\natexlab{a}}) A density-based algorithm for discovering clusters in large spatial databases with noise. In: kdd, pp 226--231

\bibitem[{Ester et~al(1996{\natexlab{b}})Ester, Kriegel, Sander, Xu et~al}]{ester1996density}
Ester M, Kriegel HP, Sander J, et~al (1996{\natexlab{b}}) A density-based algorithm for discovering clusters in large spatial databases with noise. In: kdd, pp 226--231

\bibitem[{Fortunato(2010)}]{fortunato2010community}
Fortunato S (2010) Community detection in graphs. Physics reports 486(3-5):75--174

\bibitem[{Georghiades et~al(2001)Georghiades, Belhumeur, and Kriegman}]{YaleB2001}
Georghiades AS, Belhumeur PN, Kriegman DJ (2001) From few to many: Illumination cone models for face recognition under variable lighting and pose. IEEE transactions on pattern analysis and machine intelligence 23(6):643--660

\bibitem[{Gidaris et~al(2018)Gidaris, Singh, and Komodakis}]{PretextRotation2018}
Gidaris S, Singh P, Komodakis N (2018) Unsupervised representation learning by predicting image rotations. arXiv preprint arXiv:180307728

\bibitem[{Goodfellow et~al(2014)Goodfellow, Pouget-Abadie, Mirza, Xu, Warde-Farley, Ozair, Courville, and Bengio}]{GAN2014}
Goodfellow I, Pouget-Abadie J, Mirza M, et~al (2014) Generative adversarial nets. Advances in neural information processing systems 27

\bibitem[{Gowda and Krishna(1978)}]{AggCluster1978}
Gowda KC, Krishna G (1978) Agglomerative clustering using the concept of mutual nearest neighbourhood. Pattern recognition 10(2):105--112

\bibitem[{Grill et~al(2020)Grill, Strub, Altch{\'e}, Tallec, Richemond, Buchatskaya, Doersch, Avila~Pires, Guo, Gheshlaghi~Azar et~al}]{BYOL2020}
Grill JB, Strub F, Altch{\'e} F, et~al (2020) Bootstrap your own latent-a new approach to self-supervised learning. Advances in neural information processing systems 33:21271--21284

\bibitem[{Gurumurthy et~al(2017)Gurumurthy, Kiran~Sarvadevabhatla, and Venkatesh~Babu}]{Deligan2017}
Gurumurthy S, Kiran~Sarvadevabhatla R, Venkatesh~Babu R (2017) Deligan: Generative adversarial networks for diverse and limited data. In: Proceedings of the IEEE conference on computer vision and pattern recognition, pp 166--174

\bibitem[{Hadsell et~al(2006)Hadsell, Chopra, and LeCun}]{DrLIM2006}
Hadsell R, Chopra S, LeCun Y (2006) Dimensionality reduction by learning an invariant mapping. In: 2006 IEEE computer society conference on computer vision and pattern recognition (CVPR'06), IEEE, pp 1735--1742

\bibitem[{Hastings(1970)}]{hastings1970monte}
Hastings WK (1970) Monte carlo sampling methods using markov chains and their applications

\bibitem[{He et~al(2020)He, Fan, Wu, Xie, and Girshick}]{MoCo2020}
He K, Fan H, Wu Y, et~al (2020) Momentum contrast for unsupervised visual representation learning. In: Proceedings of the IEEE/CVF conference on computer vision and pattern recognition, pp 9729--9738

\bibitem[{Hu et~al(2017)Hu, Miyato, Tokui, Matsumoto, and Sugiyama}]{IMSAT2017}
Hu W, Miyato T, Tokui S, et~al (2017) Learning discrete representations via information maximizing self-augmented training. In: International conference on machine learning, PMLR, pp 1558--1567

\bibitem[{Huang et~al(2020)Huang, Gong, and Zhu}]{PICA2020}
Huang J, Gong S, Zhu X (2020) Deep semantic clustering by partition confidence maximisation. In: Proceedings of the IEEE/CVF conference on computer vision and pattern recognition, pp 8849--8858

\bibitem[{Huang et~al(2014)Huang, Huang, Wang, and Wang}]{DEN2014}
Huang P, Huang Y, Wang W, et~al (2014) Deep embedding network for clustering. In: 2014 22nd International conference on pattern recognition, IEEE, pp 1532--1537

\bibitem[{Huang et~al(2019)Huang, Zhou, Peng, Zhang, Zhu, and Lv}]{huang2019mvscn}
Huang Z, Zhou JT, Peng X, et~al (2019) Multi-view spectral clustering network. In: Proceeings of the Twenty-Eighth International Joint Conference on Artificial Intelligence, {IJCAI-19}. International Joint Conferences on Artificial Intelligence Organization, pp 2563--2569, \doi{10.24963/ijcai.2019/356}

\bibitem[{Huang et~al(2021)Huang, Zhou, Zhu, Zhang, Lv, and Peng}]{huang2021deep}
Huang Z, Zhou JT, Zhu H, et~al (2021) Deep spectral representation learning from multi-view data. IEEE Transactions on Image Processing 30:5352--5362

\bibitem[{Huang et~al(2022)Huang, Chen, Zhang, and Shan}]{ProPos2022}
Huang Z, Chen J, Zhang J, et~al (2022) Learning representation for clustering via prototype scattering and positive sampling. IEEE Transactions on Pattern Analysis and Machine Intelligence

\bibitem[{Hubert and Arabie(1985)}]{ARI1985}
Hubert L, Arabie P (1985) Comparing partitions. Journal of classification 2:193--218

\bibitem[{Huynh et~al(2022)Huynh, Kornblith, Walter, Maire, and Khademi}]{FNC_Boosting}
Huynh T, Kornblith S, Walter MR, et~al (2022) Boosting contrastive self-supervised learning with false negative cancellation. In: Proceedings of the IEEE/CVF winter conference on applications of computer vision, pp 2785--2795

\bibitem[{Ji et~al(2019)Ji, Henriques, and Vedaldi}]{IIC2019}
Ji X, Henriques JF, Vedaldi A (2019) Invariant information clustering for unsupervised image classification and segmentation. In: Proceedings of the IEEE/CVF international conference on computer vision, pp 9865--9874

\bibitem[{Jiang et~al(2016)Jiang, Zheng, Tan, Tang, and Zhou}]{VaDE2016}
Jiang Z, Zheng Y, Tan H, et~al (2016) Variational deep embedding: An unsupervised and generative approach to clustering. arXiv preprint arXiv:161105148

\bibitem[{Jin et~al(2021)Jin, Yu, Jiao, Pan, He, Wu, Philip, and Zhang}]{jin2021survey}
Jin D, Yu Z, Jiao P, et~al (2021) A survey of community detection approaches: From statistical modeling to deep learning. IEEE Transactions on Knowledge and Data Engineering 35(2):1149--1170

\bibitem[{Kim and Ha(2021)}]{C3GAN2022}
Kim Y, Ha JW (2021) Contrastive fine-grained class clustering via generative adversarial networks

\bibitem[{Kingma and Welling(2013)}]{VAE2013}
Kingma DP, Welling M (2013) Auto-encoding variational bayes. arXiv preprint arXiv:13126114

\bibitem[{Krause et~al(2010)Krause, Perona, and Gomes}]{RIM2010}
Krause A, Perona P, Gomes R (2010) Discriminative clustering by regularized information maximization. Advances in neural information processing systems 23

\bibitem[{Krizhevsky et~al(2009)Krizhevsky, Hinton et~al}]{CIFAR2009}
Krizhevsky A, Hinton G, et~al (2009) Learning multiple layers of features from tiny images

\bibitem[{Kuhn(1955)}]{Hungarian1955}
Kuhn HW (1955) The hungarian method for the assignment problem. Naval research logistics quarterly 2(1-2):83--97

\bibitem[{Kumar et~al(2012)Kumar, Kim, and Hancke}]{kumar2012environmental}
Kumar A, Kim H, Hancke GP (2012) Environmental monitoring systems: A review. IEEE Sensors Journal 13(4):1329--1339

\bibitem[{Laine and Aila(2016)}]{laine2016temporal}
Laine S, Aila T (2016) Temporal ensembling for semi-supervised learning. arXiv preprint arXiv:161002242

\bibitem[{Le and Yang(2015)}]{TinyImageNet2015}
Le Y, Yang X (2015) Tiny imagenet visual recognition challenge. CS 231N 7(7):3

\bibitem[{Li et~al(2023{\natexlab{a}})Li, Li, Yang, Hu, Peng, and Peng}]{ProImp2023}
Li H, Li Y, Yang M, et~al (2023{\natexlab{a}}) Incomplete multi-view clustering via prototype-based imputation. arXiv preprint arXiv:230111045

\bibitem[{Li et~al(2020)Li, Zhao, and Liu}]{li2020deep}
Li P, Zhao H, Liu H (2020) Deep fair clustering for visual learning. In: Proceedings of the IEEE/CVF Conference on Computer Vision and Pattern Recognition, pp 9070--9079

\bibitem[{Li et~al(2021)Li, Hu, Liu, Peng, Zhou, and Peng}]{CC2021}
Li Y, Hu P, Liu Z, et~al (2021) Contrastive clustering. In: Proceedings of the AAAI conference on artificial intelligence, pp 8547--8555

\bibitem[{Li et~al(2022)Li, Yang, Peng, Li, Huang, and Peng}]{TCL2022}
Li Y, Yang M, Peng D, et~al (2022) Twin contrastive learning for online clustering. International Journal of Computer Vision 130(9):2205--2221

\bibitem[{Li et~al(2023{\natexlab{b}})Li, Hu, Peng, Lv, Fan, and Peng}]{TAC2023}
Li Y, Hu P, Peng D, et~al (2023{\natexlab{b}}) Image clustering with external guidance. arXiv preprint arXiv:231011989

\bibitem[{Li et~al(2023{\natexlab{c}})Li, Lin, Hu, Peng, Luo, and Peng}]{ScBatchEff2023}
Li Y, Lin Y, Hu P, et~al (2023{\natexlab{c}}) Single-cell rna-seq debiased clustering via batch effect disentanglement. IEEE Transactions on Neural Networks and Learning Systems

\bibitem[{Li et~al(2023{\natexlab{d}})Li, Zhang, Yang, Peng, Yu, Liu, Lv, Chen, and Peng}]{SCbridge2023}
Li Y, Zhang D, Yang M, et~al (2023{\natexlab{d}}) scbridge embraces cell heterogeneity in single-cell rna-seq and atac-seq data integration. Nature Communications 14(1):6045

\bibitem[{Lin et~al(2021)Lin, Gou, Liu, Li, Lv, and Peng}]{Completer2021}
Lin Y, Gou Y, Liu Z, et~al (2021) Completer: Incomplete multi-view clustering via contrastive prediction. In: Proceedings of the IEEE/CVF conference on computer vision and pattern recognition, pp 11174--11183

\bibitem[{Lin et~al(2022)Lin, Gou, Liu, Bai, Lv, and Peng}]{lin2022dcp}
Lin Y, Gou Y, Liu X, et~al (2022) Dual contrastive prediction for incomplete multi-view representation learning. IEEE Transactions on Pattern Analysis and Machine Intelligence pp 1--14. \doi{10.1109/TPAMI.2022.3197238}

\bibitem[{Lin et~al(2023)Lin, Yang, Yu, Hu, Zhang, and Peng}]{GraphM2023}
Lin Y, Yang M, Yu J, et~al (2023) Graph matching with bi-level noisy correspondence. In: Proceedings of the IEEE/CVF International Conference on Computer Vision, pp 23362--23371

\bibitem[{Liu et~al(2022)Liu, Lin, Jiang, Liu, Wen, and Peng}]{liu2022improve}
Liu J, Lin Y, Jiang L, et~al (2022) Improve interpretability of neural networks via sparse contrastive coding. In: Findings of the Association for Computational Linguistics: EMNLP 2022, pp 460--470

\bibitem[{Liu et~al(2019{\natexlab{a}})Liu, Zhu, Li, Wang, Zhu, Liu, Kloft, Shen, Yin, and Gao}]{liutong2019}
Liu X, Zhu X, Li M, et~al (2019{\natexlab{a}}) Multiple kernel $ k $ k-means with incomplete kernels. IEEE transactions on pattern analysis and machine intelligence 42(5):1191--1204

\bibitem[{Liu et~al(2019{\natexlab{b}})Liu, Zhu, Li, Wang, Zhu, Liu, Kloft, Shen, Yin, and Gao}]{liu2019multiple}
Liu X, Zhu X, Li M, et~al (2019{\natexlab{b}}) Multiple kernel $k$ k-means with incomplete kernels. IEEE transactions on pattern analysis and machine intelligence 42(5):1191--1204

\bibitem[{Lu et~al(2024)Lu, Lin, Yang, Peng, Hu, and Peng}]{DIVIDE2024}
Lu Y, Lin Y, Yang M, et~al (2024) Decoupled contrastive multi-view clustering with high-order random walks. Proceedings of the AAAI Conference on Artificial Intelligence 38(13):14193--14201

\bibitem[{MacQueen et~al(1967)}]{kmeans1967}
MacQueen J, et~al (1967) Some methods for classification and analysis of multivariate observations. In: Proceedings of the fifth Berkeley symposium on mathematical statistics and probability, Oakland, CA, USA, pp 281--297

\bibitem[{McDaid et~al(2011)McDaid, Greene, and Hurley}]{NMI2011}
McDaid AF, Greene D, Hurley N (2011) Normalized mutual information to evaluate overlapping community finding algorithms. arXiv preprint arXiv:11102515

\bibitem[{Min et~al(2018)Min, Guo, Liu, Zhang, Cui, and Long}]{SurveyOnArch2018}
Min E, Guo X, Liu Q, et~al (2018) A survey of clustering with deep learning: From the perspective of network architecture. IEEE Access 6:39501--39514

\bibitem[{Montoya et~al(2006)Montoya, Pimm, and Sol{\'e}}]{montoya2006ecological}
Montoya JM, Pimm SL, Sol{\'e} RV (2006) Ecological networks and their fragility. Nature 442(7100):259--264

\bibitem[{Moskalev et~al(2022)Moskalev, Sosnovik, Fischer, and Smeulders}]{ContraQuad2022}
Moskalev A, Sosnovik I, Fischer V, et~al (2022) Contrasting quadratic assignments for set-based representation learning. In: European Conference on Computer Vision, Springer, pp 88--104

\bibitem[{Mukherjee et~al(2019)Mukherjee, Asnani, Lin, and Kannan}]{ClusterGAN2019}
Mukherjee S, Asnani H, Lin E, et~al (2019) Clustergan: Latent space clustering in generative adversarial networks. In: Proceedings of the AAAI conference on artificial intelligence, pp 4610--4617

\bibitem[{Murdock and Yaeger(2011)}]{murdock2011identifying}
Murdock J, Yaeger LS (2011) Identifying species by genetic clustering. In: ECAL, pp 564--572

\bibitem[{Murtagh and Contreras(2012)}]{hierarchical2012}
Murtagh F, Contreras P (2012) Algorithms for hierarchical clustering: an overview. Wiley Interdisciplinary Reviews: Data Mining and Knowledge Discovery 2(1):86--97

\bibitem[{Nene et~al(1996)Nene, Nayar, Murase et~al}]{COIL20_1996}
Nene SA, Nayar SK, Murase H, et~al (1996) Columbia object image library (coil-20)

\bibitem[{Newman and Girvan(2004)}]{newman2004finding}
Newman ME, Girvan M (2004) Finding and evaluating community structure in networks. Physical review E 69(2):026113

\bibitem[{Nguyen et~al(2021)Nguyen, Bui, Duong, Bui, and Luu}]{vit_cluster2021}
Nguyen XB, Bui DT, Duong CN, et~al (2021) Clusformer: A transformer based clustering approach to unsupervised large-scale face and visual landmark recognition. In: 2021 IEEE/CVF Conference on Computer Vision and Pattern Recognition (CVPR), pp 10842--10851, \doi{10.1109/CVPR46437.2021.01070}

\bibitem[{Nie et~al(2016)Nie, Li, Li et~al}]{nie2016parameter}
Nie F, Li J, Li X, et~al (2016) Parameter-free auto-weighted multiple graph learning: a framework for multiview clustering and semi-supervised classification. In: IJCAI

\bibitem[{Nie et~al(2017)Nie, Li, Li et~al}]{mvc_graph2017}
Nie F, Li J, Li X, et~al (2017) Self-weighted multiview clustering with multiple graphs. In: IJCAI, pp 2564--2570

\bibitem[{Niu et~al(2022)Niu, Shan, and Wang}]{SPICE2022}
Niu C, Shan H, Wang G (2022) Spice: Semantic pseudo-labeling for image clustering. IEEE Transactions on Image Processing 31:7264--7278

\bibitem[{Oord et~al(2018)Oord, Li, and Vinyals}]{CPC2018}
Oord Avd, Li Y, Vinyals O (2018) Representation learning with contrastive predictive coding. arXiv preprint arXiv:180703748

\bibitem[{Peng et~al(2016)Peng, Xiao, Feng, Yau, and Yi}]{PARTY2016}
Peng X, Xiao S, Feng J, et~al (2016) Deep subspace clustering with sparsity prior. In: IJCAI, pp 1925--1931

\bibitem[{Peng et~al(2019)Peng, Huang, Lv, Zhu, and Zhou}]{COMIC2019}
Peng X, Huang Z, Lv J, et~al (2019) Comic: Multi-view clustering without parameter selection. In: International conference on machine learning, PMLR, pp 5092--5101

\bibitem[{Qian(2023)}]{SeCu2023}
Qian Q (2023) Stable cluster discrimination for deep clustering. In: Proceedings of the IEEE/CVF International Conference on Computer Vision, pp 16645--16654

\bibitem[{Radford et~al(2015)Radford, Metz, and Chintala}]{DCGAN2016}
Radford A, Metz L, Chintala S (2015) Unsupervised representation learning with deep convolutional generative adversarial networks. arXiv preprint arXiv:151106434

\bibitem[{Radford et~al(2021)Radford, Kim, Hallacy, Ramesh, Goh, Agarwal, Sastry, Askell, Mishkin, Clark et~al}]{radford2021learning}
Radford A, Kim JW, Hallacy C, et~al (2021) Learning transferable visual models from natural language supervision. In: International conference on machine learning, PMLR, pp 8748--8763

\bibitem[{Ren et~al(2022)Ren, Pu, Yang, Xu, Li, Pu, Yu, and He}]{SurveyOnData2022}
Ren Y, Pu J, Yang Z, et~al (2022) Deep clustering: A comprehensive survey. arXiv preprint arXiv:221004142

\bibitem[{Roweis and Saul(2000)}]{LLE2000}
Roweis ST, Saul LK (2000) Nonlinear dimensionality reduction by locally linear embedding. science 290(5500):2323--2326

\bibitem[{Saeedi~Emadi and Mazinani(2018)}]{anomalydetection2018}
Saeedi~Emadi H, Mazinani SM (2018) A novel anomaly detection algorithm using dbscan and svm in wireless sensor networks. Wireless Personal Communications 98:2025--2035

\bibitem[{Schaeffer(2007)}]{schaeffer2007graph}
Schaeffer SE (2007) Graph clustering. Computer science review 1(1):27--64

\bibitem[{Shah and Koltun(2017)}]{RCC2017}
Shah SA, Koltun V (2017) Robust continuous clustering. Proceedings of the National Academy of Sciences 114(37):9814--9819

\bibitem[{Shah and Koltun(2018)}]{DCC2018}
Shah SA, Koltun V (2018) Deep continuous clustering. arXiv preprint arXiv:180301449

\bibitem[{Shaham and Lederman(2018)}]{SiameseNets2018}
Shaham U, Lederman RR (2018) Learning by coincidence: Siamese networks and common variable learning. Pattern Recognition 74:52--63

\bibitem[{Shaham et~al(2018)Shaham, Stanton, Li, Nadler, Basri, and Kluger}]{SpectralNet2018}
Shaham U, Stanton K, Li H, et~al (2018) Spectralnet: Spectral clustering using deep neural networks. arXiv preprint arXiv:180101587

\bibitem[{Shen et~al(2021)Shen, Shen, Wang, Qin, Torr, and Shao}]{TCC2021}
Shen Y, Shen Z, Wang M, et~al (2021) You never cluster alone. Advances in Neural Information Processing Systems 34:27734--27746

\bibitem[{Shorten and Khoshgoftaar(2019)}]{augmentation2019}
Shorten C, Khoshgoftaar TM (2019) A survey on image data augmentation for deep learning. Journal of big data 6(1):1--48

\bibitem[{Sohn et~al(2020)Sohn, Berthelot, Li, Zhang, Carlini, Cubuk, Kurakin, Zhang, and Raffel}]{FixMatch2020}
Sohn K, Berthelot D, Li CL, et~al (2020) Fixmatch: Simplifying semi-supervised learning with consistency and confidence. arXiv preprint arXiv:200107685

\bibitem[{Song et~al(2013)Song, Liu, Huang, Wang, and Tan}]{ABDC2013}
Song C, Liu F, Huang Y, et~al (2013) Auto-encoder based data clustering. In: Progress in Pattern Recognition, Image Analysis, Computer Vision, and Applications: 18th Iberoamerican Congress, CIARP 2013, Havana, Cuba, November 20-23, 2013, Proceedings, Part I 18, Springer, pp 117--124

\bibitem[{Su et~al(2022)Su, Xue, Liu, Wu, Yang, Zhou, Hu, Paris, Nepal, Jin, Sheng, and Yu}]{communitydetection}
Su X, Xue S, Liu F, et~al (2022) A comprehensive survey on community detection with deep learning. IEEE Transactions on Neural Networks and Learning Systems pp 1--21. \doi{10.1109/TNNLS.2021.3137396}

\bibitem[{Van~Gansbeke et~al(2020)Van~Gansbeke, Vandenhende, Georgoulis, Proesmans, and Van~Gool}]{SCAN2020}
Van~Gansbeke W, Vandenhende S, Georgoulis S, et~al (2020) Scan: Learning to classify images without labels. In: European conference on computer vision, Springer, pp 268--285

\bibitem[{Wang et~al(2018)Wang, Chen, Nie, and Li}]{wang2018}
Wang Q, Chen M, Nie F, et~al (2018) Detecting coherent groups in crowd scenes by multiview clustering. IEEE transactions on pattern analysis and machine intelligence 42(1):46--58

\bibitem[{Wang et~al(2016)Wang, Yan, Lee, and Livescu}]{DCCAE2016}
Wang W, Yan X, Lee H, et~al (2016) Deep variational canonical correlation analysis. arXiv preprint arXiv:161003454

\bibitem[{Wang et~al(2021)Wang, Ni, Jing, Wang, Zhang, and Xing}]{wang2021dnb}
Wang Z, Ni Y, Jing B, et~al (2021) Dnb: A joint learning framework for deep bayesian nonparametric clustering. IEEE Transactions on Neural Networks and Learning Systems 33(12):7610--7620

\bibitem[{Wright et~al(2010)Wright, Ma, Mairal, Sapiro, Huang, and Yan}]{SparseRepresentaion2010}
Wright J, Ma Y, Mairal J, et~al (2010) Sparse representation for computer vision and pattern recognition. Proceedings of the IEEE 98(6):1031--1044

\bibitem[{Wu et~al(2019)Wu, Zheng, Zhang, Yuan, Cheng, Zhao, Lin, Zhao, Jiang, and Huang}]{wu2019deep}
Wu D, Zheng SJ, Zhang XP, et~al (2019) Deep learning-based methods for person re-identification: A comprehensive review. Neurocomputing 337:354--371

\bibitem[{Wu et~al(2016)Wu, Tan, and Xiong}]{wu2016data}
Wu M, Tan L, Xiong N (2016) Data prediction, compression, and recovery in clustered wireless sensor networks for environmental monitoring applications. Information Sciences 329:800--818

\bibitem[{Wu et~al(2018)Wu, Xiong, Yu, and Lin}]{PretextInstDist2018}
Wu Z, Xiong Y, Yu SX, et~al (2018) Unsupervised feature learning via non-parametric instance discrimination. In: Proceedings of the IEEE conference on computer vision and pattern recognition, pp 3733--3742

\bibitem[{Xia and Vlajic(2007)}]{xia2007near}
Xia D, Vlajic N (2007) Near-optimal node clustering in wireless sensor networks for environment monitoring. In: 21st international conference on advanced information networking and applications (AINA'07), IEEE, pp 632--641

\bibitem[{Xie et~al(2016)Xie, Girshick, and Farhadi}]{DEC2016}
Xie J, Girshick R, Farhadi A (2016) Unsupervised deep embedding for clustering analysis. In: International conference on machine learning, PMLR, pp 478--487

\bibitem[{Xu et~al(2013)Xu, Tao, and Xu}]{xu2013}
Xu C, Tao D, Xu C (2013) A survey on multi-view learning. arXiv preprint arXiv:13045634

\bibitem[{Xu et~al(2022)Xu, De~Mello, Liu, Byeon, Breuel, Kautz, and Wang}]{groupvit2022}
Xu J, De~Mello S, Liu S, et~al (2022) Groupvit: Semantic segmentation emerges from text supervision. In: Proceedings of the IEEE/CVF Conference on Computer Vision and Pattern Recognition, pp 18134--18144

\bibitem[{Yan et~al(2023)Yan, Li, Qin, Zheng, Liao, and Yang}]{yan2023efficient}
Yan Y, Li J, Qin J, et~al (2023) Efficient person search: An anchor-free approach. International Journal of Computer Vision pp 1--20

\bibitem[{Yang et~al(2016)Yang, Parikh, and Batra}]{JULE2016}
Yang J, Parikh D, Batra D (2016) Joint unsupervised learning of deep representations and image clusters. In: Proceedings of the IEEE conference on computer vision and pattern recognition, pp 5147--5156

\bibitem[{Yang et~al(2023)Yang, Liu, Xu, and Huang}]{uda2023}
Yang J, Liu J, Xu N, et~al (2023) Tvt: Transferable vision transformer for unsupervised domain adaptation. In: Proceedings of the IEEE/CVF Winter Conference on Applications of Computer Vision, pp 520--530

\bibitem[{Yang et~al(2021)Yang, Li, Huang, Liu, Hu, and Peng}]{yang2021MvCLN}
Yang M, Li Y, Huang Z, et~al (2021) Partially view-aligned representation learning with noise-robust contrastive loss. In: Proceedings of the IEEE/CVF Conference on Computer Vision and Pattern Recognition (CVPR)

\bibitem[{Yang et~al(2022{\natexlab{a}})Yang, Huang, Hu, Li, Lv, and Peng}]{yang2022DART}
Yang M, Huang Z, Hu P, et~al (2022{\natexlab{a}}) Learning with twin noisy labels for visible-infrared person re-identification. In: Proc. IEEE Conf. Comput. Vis. Pattern Recognit.

\bibitem[{Yang et~al(2022{\natexlab{b}})Yang, Li, Hu, Bai, Lv, and Peng}]{SURE2022}
Yang M, Li Y, Hu P, et~al (2022{\natexlab{b}}) Robust multi-view clustering with incomplete information. IEEE Trans Pattern Anal Mach Intell

\bibitem[{Ye et~al(2021)Ye, Shen, Lin, Xiang, Shao, and Hoi}]{reid2021}
Ye M, Shen J, Lin G, et~al (2021) Deep learning for person re-identification: A survey and outlook. IEEE transactions on pattern analysis and machine intelligence 44(6):2872--2893

\bibitem[{Zeng et~al(2023)Zeng, Li, Hu, Peng, Lv, and Peng}]{zeng2023deep}
Zeng P, Li Y, Hu P, et~al (2023) Deep fair clustering via maximizing and minimizing mutual information: Theory, algorithm and metric. In: Proceedings of the IEEE/CVF Conference on Computer Vision and Pattern Recognition, pp 23986--23995

\bibitem[{Zhang et~al(2015)Zhang, Fu, Liu, Liu, and Cao}]{zhang2015}
Zhang C, Fu H, Liu S, et~al (2015) Low-rank tensor constrained multiview subspace clustering. In: Proceedings of the IEEE international conference on computer vision, pp 1582--1590

\bibitem[{Zhang et~al(2023)Zhang, Nie, and Li}]{zhang2023large}
Zhang H, Nie F, Li X (2023) Large-scale clustering with structured optimal bipartite graph. IEEE Transactions on Pattern Analysis and Machine Intelligence

\bibitem[{Zhang et~al(2019)Zhang, Qi, Wang, and Luo}]{PretextTrans2019}
Zhang L, Qi GJ, Wang L, et~al (2019) Aet vs. aed: Unsupervised representation learning by auto-encoding transformations rather than data. In: Proceedings of the IEEE/CVF Conference on Computer Vision and Pattern Recognition, pp 2547--2555

\bibitem[{Zhang et~al(2018)Zhang, Liu, Shen, Shen, and Shao}]{BMVC2018}
Zhang Z, Liu L, Shen F, et~al (2018) Binary multi-view clustering. IEEE transactions on pattern analysis and machine intelligence 41(7):1774--1782

\bibitem[{Zhao et~al(2016)Zhao, Liu, and Fu}]{MKK}
Zhao H, Liu H, Fu Y (2016) Incomplete multi-modal visual data grouping. In: IJCAI, pp 2392--2398

\bibitem[{Zhao et~al(2019)Zhao, Wang, Masoomi, and Dy}]{zhao2019streaming}
Zhao T, Wang Z, Masoomi A, et~al (2019) Streaming adaptive nonparametric variational autoencoder. arXiv preprint arXiv:190603288

\bibitem[{Zhong et~al(2020)Zhong, Chen, Jin, and Hua}]{DRC2020}
Zhong H, Chen C, Jin Z, et~al (2020) Deep robust clustering by contrastive learning. arXiv preprint arXiv:200803030

\bibitem[{Zhong et~al(2021)Zhong, Wu, Chen, Huang, Deng, Nie, Lin, and Hua}]{GCC2021}
Zhong H, Wu J, Chen C, et~al (2021) Graph contrastive clustering. In: Proceedings of the IEEE/CVF international conference on computer vision, pp 9224--9233

\bibitem[{Zhou et~al(2022)Zhou, Xu, Zheng, Chen, Bu, Wu, Wang, Zhu, Ester et~al}]{SurveyOnTrain2022}
Zhou S, Xu H, Zheng Z, et~al (2022) A comprehensive survey on deep clustering: Taxonomy, challenges, and future directions. arXiv preprint arXiv:220607579

\end{thebibliography}
%% if required, the content of .bbl file can be included here once bbl is generated
%%\input sn-article.bbl

\section*{Author Contributions}
All authors contributed to the core insights presented in this paper. Xi Peng supervised this survey and provided valuable guidance throughout the process. Yiding Lu, Haobin Li, Yunfan Li, and Yijie Lin collaboratively wrote \textbf{Priors for Deep Clustering}. Yiding Lu took the lead in crafting \textbf{Introduction}, \textbf{Application}, and \textbf{Future Challenges}. Haobin Li was responsible for collecting and analyzing experimental results, creating figures, and summarizing tables. Yunfan Li and Yijie Lin designed the outline, wrote \textbf{Abstract}, and refined the manuscript.

\section*{Data Availability}
The datasets utilized in this survey are publicly available and can be accessed from the following sources:

\begin{itemize}
    \item CIFAR-10 and CIFAR-100:\\
    \url{https://www.cs.toronto.edu/~kriz/cifar.html}.
    \item STL-10:\\
    \url{https://cs.stanford.edu/~acoates/stl10/}.
    \item ImageNet-10 and ImageNet-Dogs: \\ \href{https://drive.google.com/drive/folders/1RGB0YxLpFlq8KXdrmHtAjdoYgxKq5-i5}{Google Drive} (Preprocessed versions)
    \item Tiny-ImageNet: \url{http://cs231n.stanford.edu/tiny-imagenet-200.zip}.
    \item ImageNet-1K: \url{https://www.image-net.org/}.
\end{itemize}

\end{document}